\documentclass{article}
\usepackage{arxiv}
\usepackage{graphicx,amsfonts,amssymb,amsmath,mathrsfs,hyperref,bm,dsfont}
\usepackage[normalem]{ulem}
\usepackage{amsfonts}  
\usepackage{amsmath}
\usepackage{mathtools}
\usepackage{xcolor}         
\usepackage{array} 
\newcolumntype{H}{>{\iffalse}c<{\fi}@{}}
\usepackage{lipsum}
\usepackage{amssymb}
\usepackage{multicol}
\usepackage{graphicx}
\usepackage{bm}
\usepackage{amsthm}
\usepackage{amsmath}
\usepackage{algorithm}
\usepackage{algpseudocode}
\usepackage{amsthm,amssymb,amsmath,bbm}
\renewcommand{\P}{\mathbb{P}}
\newcommand{\E}{\mathbb{E}}
\newcommand{\RR}{\mathbb{R}}

\newcommand{\F}{\mathcal{F}}
\renewcommand{\O}{\Omega}

\usepackage[numbers]{natbib}
\usepackage{authblk}

\usepackage{color}

\newcommand\given[1][]{\:#1\vert\:}

\newif\ifhyper
\hypertrue
\ifhyper
\hypersetup{
   citecolor = {red},
   colorlinks = {true}, 
   linkcolor = {blue},
   urlcolor = {blue} 
}
\fi

\def\be{\begin{equation}}
\def\ee{\end{equation}}
\def\bea{\begin{eqnarray}}
\def\eea{\end{eqnarray}}

\title{Quantum-Inspired Tensor Neural Networks for Partial Differential Equations}

\author[1,4]{\thanks{Email: raj.patel@multiversecomputing.com}\hspace{0.5mm} Raj G. Patel}
\author[2]{Chia-Wei Hsing}
\author[2]{Serkan Şahin}
\author[2,5]{Saeed S. Jahromi}
\author[1]{Samuel Palmer}
\author[2]{Shivam Sharma}
\author[3]{Christophe Michel}
\author[3]{Vincent Porte}
\author[3]{Mustafa Abid}
\author[3]{St\'ephane Aubert}
\author[3]{Pierre Castellani}
\author[4]{Chi-Guhn Lee}
\author[1]{Samuel Mugel}
\author[2,5,6]{Rom\'an Orús}

\affil[1]{%
Multiverse Computing, Centre for Social Innovation, 192 Spadina Ave, Suite 509, Toronto, M5T 2C2, Canada} 
\affil[2]{%
Multiverse Computing, Paseo de Miram\'on 170, 20014 San Sebasti\'an, Spain} 
\affil[3]{%
Cr\'edit Agricole, 12, Place des Etats-Unis - CS 70052 - 92547 Montrouge Cedex, France} 
\affil[4]{%
University of Toronto, Toronto, Ontario M5S 2E4, Canada} 
\affil[5]{%
Donostia International Physics Center, Paseo Manuel de Lardizabal 4, E-20018 San Sebasti\'an, Spain}
\affil[6]{%
Ikerbasque Foundation for Science, Maria Diaz de Haro 3, E-48013 Bilbao, Spain}

\begin{document}
\maketitle
\begin{abstract}
	Partial Differential Equations (PDEs) are used to model a variety of dynamical systems in science and engineering. Recent advances in deep learning have enabled us to solve them in a higher dimension by addressing the curse of dimensionality in new ways. However, deep learning methods are constrained by training time and memory. To tackle these shortcomings, we implement Tensor Neural Networks (TNN), a quantum-inspired neural network architecture that leverages Tensor Network ideas to improve upon deep learning approaches. We demonstrate that TNN provide significant parameter savings while attaining the same accuracy as compared to the classical Dense Neural Network (DNN). In addition, we also show how TNN can be trained faster than DNN for the same accuracy. We benchmark TNN by applying them to solve parabolic PDEs, specifically the Black-Scholes-Barenblatt equation, widely used in financial pricing theory, empirically showing the advantages of TNN over DNN. Further examples, such as the Hamilton-Jacobi-Bellman equation, are also discussed.
\end{abstract}

\section{Introduction}

\noindent
Partial Differential Equations (PDEs) are an indispensable tool for modeling a variety of problems ranging from physics to finance. Typical approaches for solving such PDEs mostly rely on classical mesh-based numerical methods or Monte Carlo approaches. However, scaling these to higher dimensions has always been a challenge because of their dependency on spatio-temporal grids as well as on a large number of sample paths. As an alternative, recent advancements in deep learning have enabled us to circumvent some of these challenges by approximating the unknown solution using Dense Neural Networks (DNNs) \cite{Raissi,Beck_2019,Han_2018, E_2017}. The basic idea of these approaches is to leverage the connection between high-dimensional PDEs and forward-backward stochastic differential equations (FBSDE) \cite{Cheridito}. The solution of the corresponding FBSDE can be written as a deterministic function of time and the state process. Under suitable regularity assumptions, the FBSDE solution can represent the solution of the underlying parabolic PDE. Efficient methods for approximating the FBSDE solution with a DNN have been put forward recently in Refs. \cite{Raissi-part1,Raissi-part2}. However, in spite of their apparent success, DNN approaches for solving PDEs are computationally expensive and limited by memory \cite{TNN_NIPS, tn_memory, xue13_interspeech}. \newline

\noindent
In this paper we show how to overcome this problem by combining Tensor Networks (TN) \cite{RomanTN} with DNNs. TNs were proposed in physics to efficiently describe strongly-correlated structures, such as quantum many-body states of matter. They are at the basis of well-known numerical simulation methods, such as Density Matrix Renormalization Group (DMRG) \citep{dmrg}, Time-Evolving Block Decimation (TEBD) \cite{tebd}, and more. At the fundamental level, TNs are nothing but efficient descriptions of high-dimensional vectors and operators. Because of this, TNs have recently found important applications also in different domains of computer science such as machine learning (ML) and optimization. In particular, TNs have proven successful in ML tasks such as classification \cite{NIPS2016_6211,Stoudenmire_2018,glasser2018supervised,efthymiou2019tensornetwork,bhatia2019matrix,Liu_2019,9058650}, generative modeling \cite{PhysRevX.8.031012,PhysRevB.99.155131,PhysRevB.101.075135} and sequence modeling \cite{bradley2020modeling}. \newline

\noindent
In practice, we use a specific type of TN known as Matrix Product Operator (MPO) (also known as Tensor Train as used in Ref. \cite{TNN_NIPS}), which generalizes the low-rank idea by representing the weight matrix in terms of an efficient TN description. Following ideas from Refs. \cite{TNN_NIPS, tnn_osedelets}, we transform a DNN into what we call a Tensor Neural Network (TNN), in turn enhancing training performance and reducing memory consumption. Our approach saves memory since it uses fewer parameters compared to a standard Neural Network (NN) \cite{TNN_NIPS}. While low-rank representations of the weight matrices had already been discussed in the literature \cite{low_rank, xue13_interspeech, NIPS2013_7fec306d}, we certify the improvement obtained by TNN by showing that we cannot find an equally-performing DNN with the same number of parameters as the TNN. We achieve this by analyzing the entire sample space of DNN architectures with equivalent number of parameters as that of a TNN. To the best of our knowledge, this important milestone has not yet been considered in the literature. Our main test bed for benchmark is the Black-Scholes-Barenblatt equation, widely used in financial pricing theory. \newline

\noindent
This paper is organized as follows: in Section \ref{sec:nn-math}, we show how a parabolic PDE can essentially be mapped to an SDE and how the corresponding SDE can be solved with a neural network. In Section \ref{sec:tn}, we briefly review the concept of TN and show how one can incorporate them in NN to obtain TNN. In Section \ref{sec:example} we demonstrate how TNN can outperform DNN for the case of Black-Scholes-Barenblatt PDE. Finally, in Section \ref{sec:conclude} we present our conclusions. 
Furthermore, Appendix \ref{F-K} discusses mathematical details of the Feynman-Kac representation, and Appendix \ref{appendix:experiments} shows further experiments, including benchmarks for the Hamilton-Jacobi-Bellman equation.

\section{Neural Networks for PDE} \label{NN-Math}
\label{sec:nn-math}

The connection between PDE and Forward Backward Stochastic Differential Equations (FBSDE) is well studied in Refs. \cite{Cheridito, Antonelli, 4-stepPDE, quasi-PDE, quasi-FBPDE}. For $t \in [0,T]$ consider the following system of Stochastic Differential Equations (SDE),
\vspace{-0.15cm}
\begin{equation} \label{sde}
    \begin{split}
     dX_t &= \mu(t,X_t,Y_t,Z_t)dt + \sigma(t,X_t,Y_t)dW_t \\
     dY_t &= \varphi(t,X_t,Y_t,Z_t)dt + Z_t ^T \sigma(t,X_t,Y_t)dW_t,
    \end{split}
\end{equation}
\noindent
where $X_0 = \epsilon$ is the initial condition for the forward SDE whereas $Y_T = g(X_T)$ is the terminal condition for the backward SDE with known $g:\mathbb{R}^d \rightarrow \mathbb{R}$. Furthermore, $W_t$ is the vector valued Brownian motion. A solution to this set of equations would have $X_t, Y_t, Z_t$. 

\noindent
Now, consider the non-linear PDE
\vspace{-0.15cm}
\begin{equation} \label{pde}
    u_t = f(t,x,u,\mathcal{D}u,\mathcal{D}^2 u),
\end{equation}

\noindent
where $\mathcal{D}u, \mathcal{D}^2 u$ represents gradient and Hessian of $u$ respectively whereas function $f$ is given by
\vspace{-0.6cm}

\begin{equation}
    f(t,x,y,z,\gamma) = \varphi(t,x,y,z) - \mu(t,x,y,z)^T z - \frac{1}{2}{\rm Tr}\left(\sigma(t,x,y)\sigma(t,x,y)^T \gamma\right). 
\end{equation}

\noindent
For the given function $f$ and terminal condition $u(T, x) = g(x)$, it is known that $Y_t=u(t,X_t)$ and $Z_t=\nabla u(t,X_t)$ by Ito's Lemma. This is how the SDEs in  Eq.(\ref{sde}) are related to the PDE in Eq.(\ref{pde}), so that if we know the solution to one, we can find or approximate the other.\newline

\noindent
Now, if the solution to the PDE in Eq.(\ref{pde}) is known, then we can easily approximate the solution to SDEs in Eq.(\ref{sde}) by leveraging Euler-Maruyama discretization of SDEs in Eq.(\ref{sde}) such that:
\vspace{-0.15cm}
\begin{equation} \label{discretization}
    \begin{split}
    \Delta W_n &\sim \mathcal{N}(0,\Delta t_n) \\
    X_{n+1} - X_n &\approx \mu(t_n,X_n,Y_n,Z_n)\Delta t_n + \sigma(t_n,X_n,Y_n,Z_n) \Delta W_n \\
    \centering
    Y_{n+1} - Y_n &\approx \varphi(t_n,X_n,Y_n,Z_n)\Delta t_n + Z_n ^T \sigma(t_n,X_n,Y_n,Z_n) \Delta W_n. \\
    \end{split}
\end{equation}

\noindent
where $Y_n = u(t_n, X_n)$ and $Z_n = \nabla u(t_n, X_n)$. However, as the solution of the PDE is not known, we parameterize the solution $u_t$ using a neural network with parameters $\theta$ as proposed in Ref. \cite{Raissi}. We obtain the required gradient
vector using automatic differentiation. Parameters of the neural network representing $u(t, x)$ can be learned by minimizing the following loss function obtained by discretizing the SDE as done in Eq.(\ref{discretization}): 
\begin{multline}
\underset{\theta}{\rm min}\Bigg( \sum_{m=1}^M \sum_{n=0}^{N-1} \big \lvert Y_{n+1}^m (\theta) - Y_{n}^m (\theta) - \varphi(t_n, X_n ^m, Y_n ^m(\theta),Z_n ^m(\theta)) \Delta t_n  \\ 
- (Z_n ^m (\theta))^T \sigma(t_n, X_n ^m, Y_n ^m(\theta)) \Delta W_n ^m \big \rvert ^2+ \sum_{m=1}^M (Y_N ^m(\theta) - g(X_N ^m))^2\Bigg).
\end{multline}

\noindent
where $M$ is the batch size, $m$ labels the $m$-th realization of the underlying Brownian motion, and $n$ labels time $t_n$.\newline

\noindent
The goal is to find the solution of a given PDE, and we aim to approximate it using a neural network. In the process of doing so, we first have the PDE alongside an SDE of an underlying state with known initial condition. From this, we can get the SDE of the state variable modeled by the PDE using Ito's Lemma. This SDE has a known terminal condition. Once we have a system of FBSDE, we can use the approach described here to find or approximate the solution of the SDE which, in turn, can be used as an approximation for the solution of the PDE we are interested in.

\section{Tensor Neural Networks}
\label{sec:tn}

\subsection{Tensor Networks}

\begin{figure}[htp!]
\centerline{\includegraphics[scale=0.6]{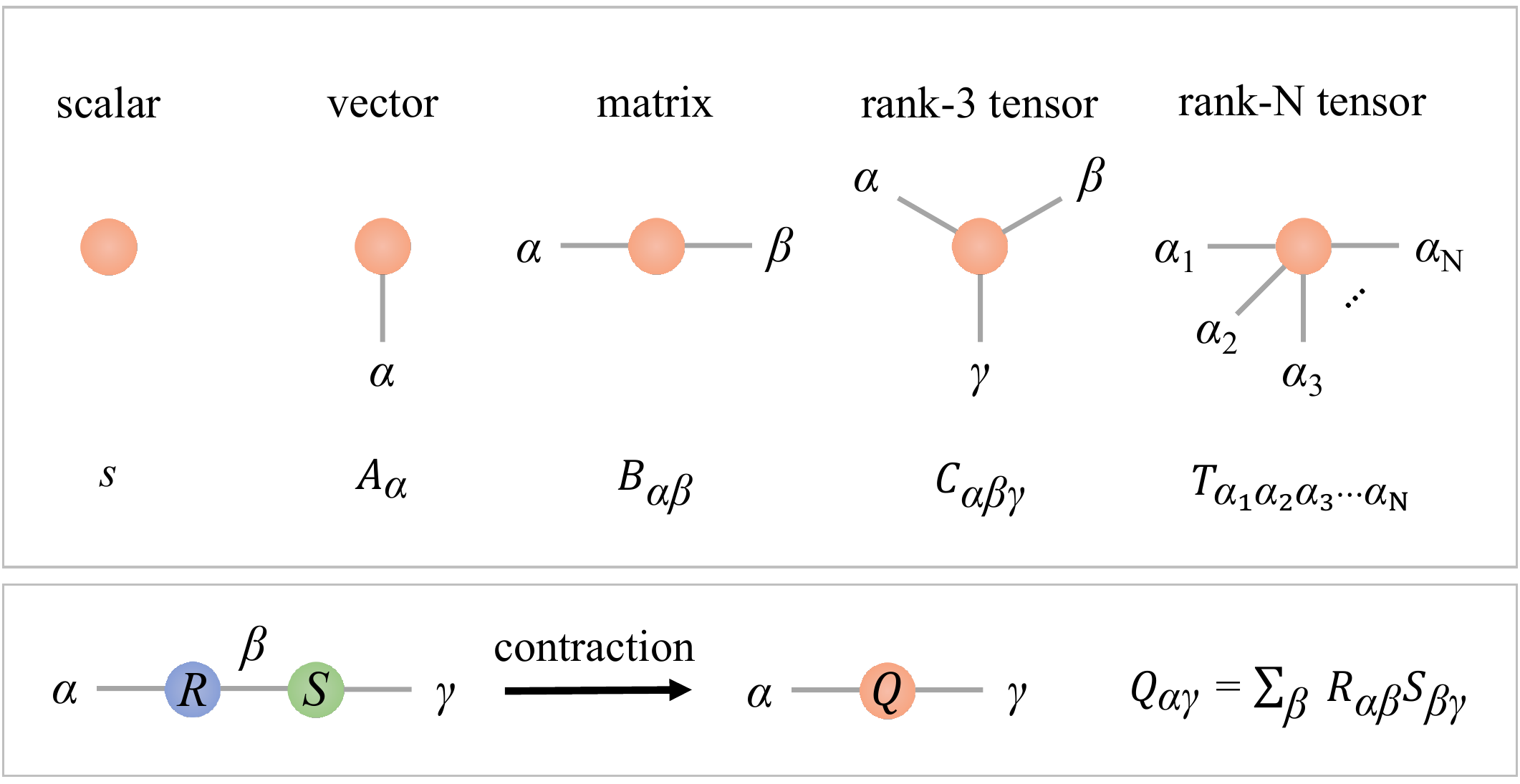}}
\caption{(\textbf{upper panel}) Tensor network diagrams. Each rank-$n$ tensor is represented by a shape with $n$ legs attached, each of which represents an index of the tensor. A scalar, vector, matrix and rank-$N$ tensor have $0$, $1$, $2$, $N$ legs attached, respectively. (\textbf{lower panel}) Contraction of tensors equivalent to matrix multiplication. Here, $R$ and $S$ tensors are connected with the shared leg $\beta$. The resulting tensor (matrix) $Q$ after full contraction has only open legs left.}
\label{Fig:Fig1}
\end{figure}

\noindent
For our purposes, a tensor is a multi-dimensional array of real/complex numbers, denoted as $T_{\alpha_{1}\alpha_{2}\alpha_{3}\ldots}$, where the indices denote different array dimensions, with the rank of the tensor defined as the number of indices. Tensor network diagrams were introduced as a convenient notation for dealing with the calculation of multiple tensors with multiple indices. As illustrated in Fig.~\ref{Fig:Fig1}, a rank-$n$ tensor is an object with $n$ legs corresponding to its $n$ indices, with each index corresponding to a dimension of the array. \newline

\begin{figure}[htp!]
\centerline{\includegraphics[scale=0.6]{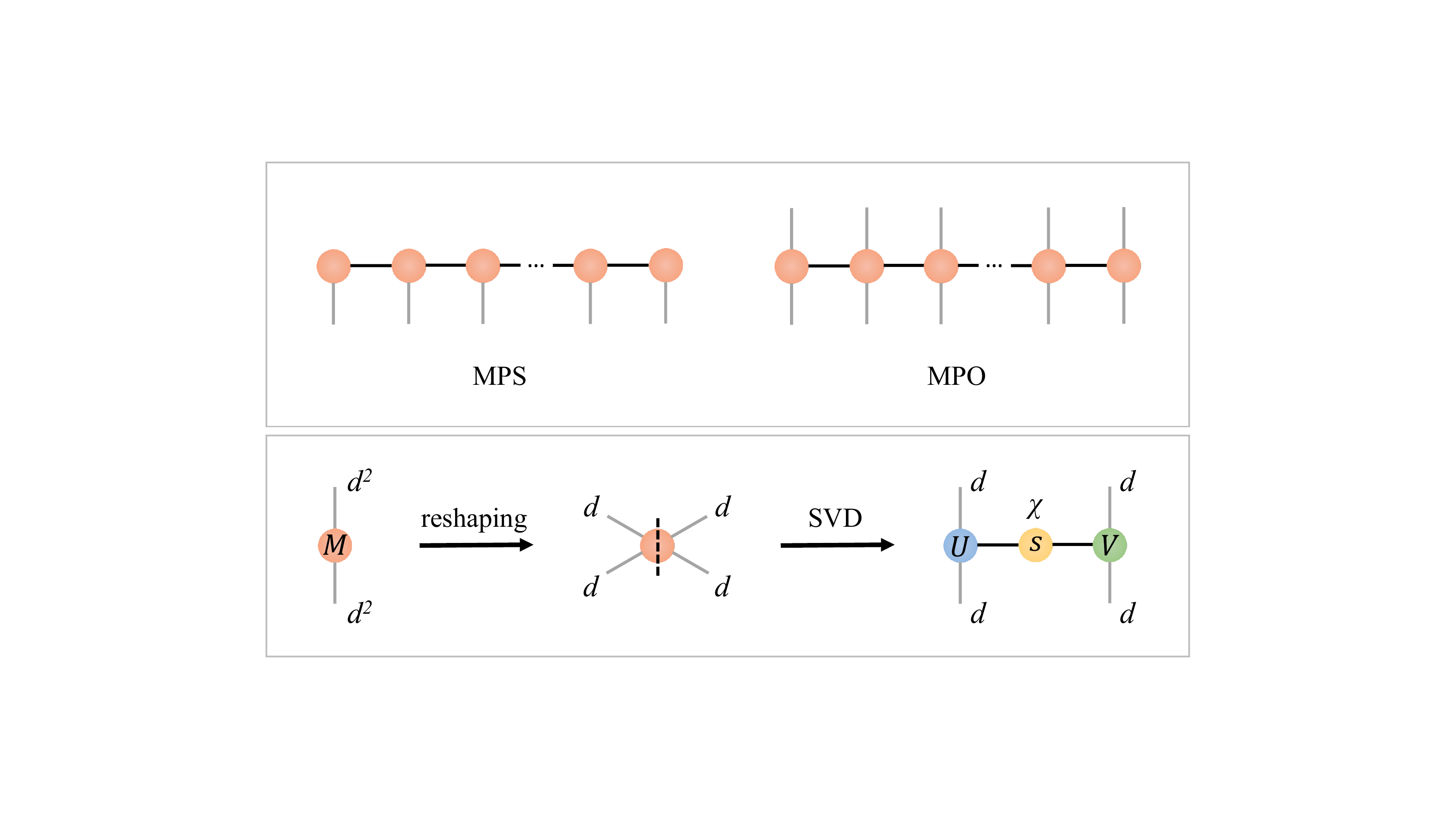}}
\caption{(\textbf{upper panel}) Matrix Product State (MPS) and Matrix Product Operator (MPO). (\textbf{lower panel}) Tensorizing a matrix into a 2-node MPO by SVD decomposition, producing two isometric tensors $U$ and $V$, as well as a diagonal matrix $S$ of real and positive singular values.}
\label{Fig:Fig2}
\end{figure}

\noindent
A tensor network is a network of interconnected tensors. By definition, a connected leg between neighboring tensors indicates a shared index to be summed over. The operation of summing over shared indices is also called tensor contraction, and a full contraction will leave a TN with only open (non-contracted) legs. For example, contraction of two rank-$2$ tensors, i.e., matrices $R_{\alpha\beta}$ and $S_{\beta\gamma}$ along the index $\beta$ is equivalent to a matrix multiplication. Diagrammatically, the pre-contraction state is shown by connecting the two tensors with their $\beta$ legs, as depicted in the lower panel of Fig.~\ref{Fig:Fig1}. Typical one-dimensional TNs include Matrix Product States (MPS) and Matrix Product Operators (MPO) \cite{RomanTN} as illustrated in Fig.~\ref{Fig:Fig2}. \newline

\noindent
Generally speaking, MPOs are efficient representations of linear operators, i.e., matrices. In general, a $d^n \times d^n$ matrix can be decomposed (tensorized) into a $n$-site MPO by a series of consecutive singular value decompositions (SVD). In such a decomposition, at every step there exist at most $d^2$ singular values. Notably, by discarding sufficiently small singular values we can find an approximated but more efficient MPO representation of the matrix. The lower panel of Fig.~\ref{Fig:Fig2} shows an example of tensorizing a $d^2 \times d^2$ matrix into a 2-site MPO. The grey legs denote what we call the \emph{physical indices} (i.e., those that correspond to the original tensor), whereas the black line in-between represents what we call a \emph{virtual bond}, with bond dimension $\chi$, corresponding to the number of singular values. 

\subsection{Tensorizing Neural Networks}

\begin{figure}[!htp]
\centering
\includegraphics[scale=0.6]{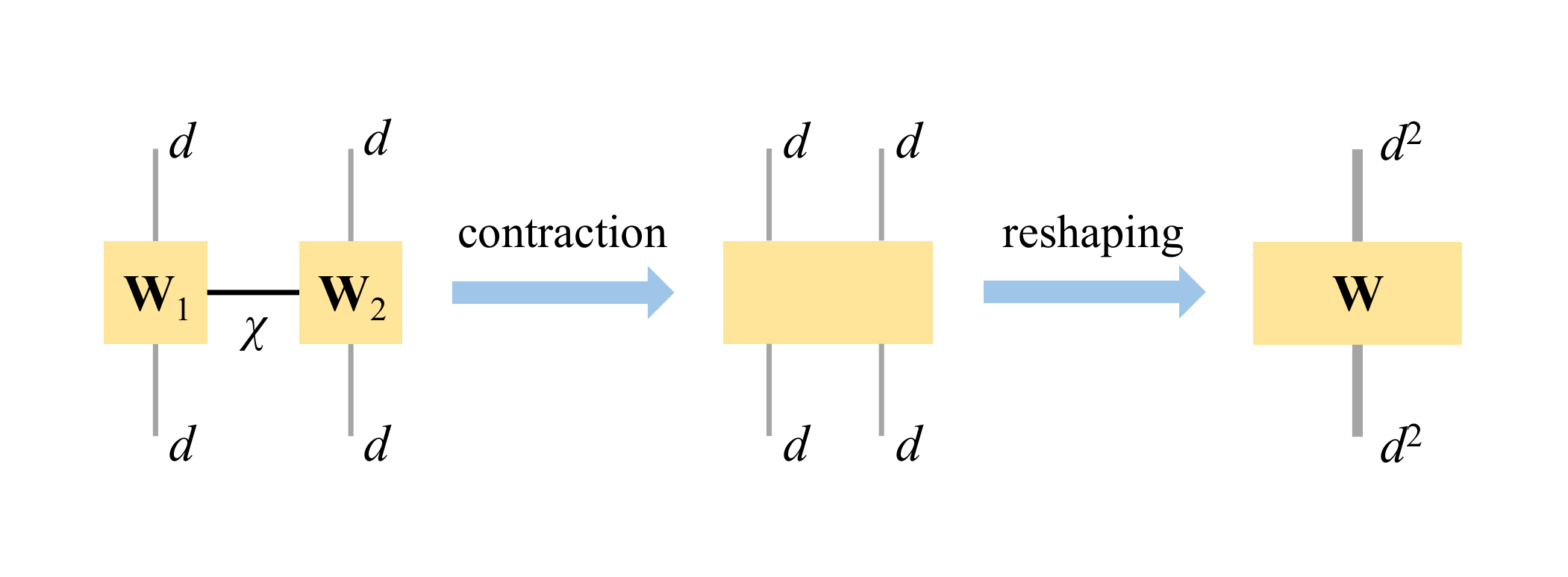}
\caption{The process of contracting a 2-node MPO and reshaping it into the weight matrix $\mathbf{W}$ in each forward pass.}
\label{Fig:Fig3}
\end{figure}

\noindent
A way of tensorizing Neural Networks is to replace the weight matrix of a dense layer by a TN. In particular, we choose an MPO representation of the weight matrix that is analogous to the Tensor-Train format \cite{TNN_NIPS}, and we call this layer a \emph{TN layer}. This representation, however, is not unique, and is determined by two additional parameters: the MPO bond dimension, and the number of tensors in the MPO. In the simplest case, the MPO may consist of just two tensors only, $\mathbf{W_1}$ and $\mathbf{W_2}$, as shown in Fig.~\ref{Fig:Fig3}. The MPO in the figure has bond dimension $\chi$ and physical dimensions $d$ everywhere. The TN layer with such an MPO can be initialized in the same manner as a weight matrix of a dense layer.\newline

\noindent
The forward pass of the TN layer involves additional operations compared to the one for a dense layer. Here, \emph{we first contract the MPO along the bond index} and then reshape the resulting rank-4 tensor into a matrix as shown in Fig.~\ref{Fig:Fig3}. This matrix is the corresponding weight matrix of a TN layer. The weight matrix can then be multiplied with the input vector. We apply an activation function to the resulting output vector, thereby finishing the forward pass of the TN layer. The weight matrix takes the form 
\begin{equation}
    \mathbf{W} = \sum_{\alpha = 1}^{\chi} (\mathbf{A}_\alpha \otimes \mathbf{B}_\alpha), ~~~\mathbf{W} \in \mathbb{R}^{d^2 \times d^2},
\label{sum_of_tensor_product}
\end{equation}

\noindent
where $\mathbf{W}_1 = [\mathbf{A}_1, \mathbf{A}_2, \cdots, \mathbf{A}_{\chi}], \mathbf{A}_\alpha \in \mathbb{R}^{d \times d}$ and $\mathbf{W}_2 = [\mathbf{B}_1, \mathbf{B}_2, \cdots, \mathbf{B}_{\chi}], \mathbf{B}_\alpha \in \mathbb{R}^{d \times d}$ are the two rank-3 weight tensors connected by a virtual bond $\alpha$ of dimension $\chi$. The resulting weight matrix $\mathbf{W}$ is of dimension $d^2 \times d^2$, so it contains $d^4$ elements. Notice that these elements \emph{are not independent} since they come from the TN structure with $2 \chi d^2$ trainable parameters. So, if we initialized the MPO with bond dimension $\chi = d^2/2$, we would have the same number of parameters as a dense layer with $d^2$ neurons. Any choice where $\chi < d^2 / 2$ will result in a weight matrix $\mathbf{W}$ comprising of $d^4 - 2 \chi d^2$ fewer parameters than the weight matrix of a dense layer, thus allowing for potential parameter savings. In principle, when $\chi = d^2$, we have sufficient degree of freedom to be able to construct an arbitrary $d^2 \times d^2$ matrix. Thus, we expect that by increasing the bond dimension the TN layer behaves increasingly similar to a dense layer. This is also shown empirically in Section \ref{sec:example}.\newline

\noindent
The existence of Kronecker product in Eq.(\ref{sum_of_tensor_product}) implies that there is correlation between the matrix elements in $\mathbf{W}$, i.e. each element will be a sum of products of elements of the tensors $\mathbf{A}$ and $\mathbf{B}$. The parameters to be trained are not the matrix elements of the weight matrix, but the elements of the individual tensors of the MPO. This can exhibit interesting training behavior and can lead to faster convergence of the loss function as we show in Section \ref{sec:example}.\newline

\noindent
By implementing the TN layer in this way and with a ML library which supports automatic differentiation such as TensorFlow or PyTorch, one can optimize the MPO weights in a similar fashion as those of dense layers in DNN and train the TNN. As an alternative, one could work with an explicit TN layer without contraction of the MPO, including tensor optimizations as in other TN algorithms (e.g., variational), provided one can find a way to decompose the activation function. We observe, however, that for most interesting NN structures, we do not actually need this option. Additionally, the TNN structure is not limited only to a single TN layer but can further be extended to any desired number of TN layers or a combination of dense and TN layers. This provides a  flexibility of designing TNN architectures which is favorable for the problems of interest.\newline

\section{Black-Scholes-Barenblatt Equation in 10 dimensions}
\label{sec:example}

We test our approach on an application in mathematical finance where we approximate the solution of a 10-dimensional Black-Scholes-Barenblatt equation to price a European-style option. This is primarily motivated by the pioneering work of using deep learning to solve high-dimensional PDEs \cite{Raissi, Beck_2019, Han_2018}. \newline

\noindent
We aim to solve the following PDE: 
\begin{equation}
    u_t = -\frac{1}{2}{\rm Tr} \left(\sigma^2 {\rm diag}(X_t ^2)\mathcal{D}^2 u \right) + r(u-(\mathcal{D}u)'x),
\end{equation}

\noindent
with terminal condition $u(T,x) = \| x \|^2$. The explicit solution for this equation is
\begin{equation}
    u(t,x) = {\rm exp}((r + \sigma^2)(T-t))\|x\|^2, 
\end{equation}
which can be used to test the accuracy of the proposed algorithm.  Using the approaches from Section \ref{NN-Math} and the Feynman-Kac formalism from Appendix \ref{F-K}, this can be re-casted into a system of forward-backward stochastic differential equations:
\begin{align}
\begin{split}
dX_t &= \sigma {\rm diag}(X_t)dW_t, ~~~ t \in [0,T] \\
X_0 &= \epsilon \\
dY_t &= r(Y_t - Z_t 'X_t)dt + \sigma Z_t ' {\rm diag}(X_t)dW_t, ~~~ t \in[0,T)\\
Y_T &= ||X_T||^2\\
\end{split}
\end{align}

\noindent 
where $\epsilon = (1,1,1, \dots, 1) \in \mathbb{R}^{10}$, $T=1$, $\sigma=0.4$, $r=0.05$. $W_t$ is a vector-valued Brownian motion, whereas $Y_t$ and $Z_t$ represent $u(t,X_t)$ and $Du(t,X_t)$. Furthermore, we partition the time domain $[0, T]$ into $N = 50$ equally spaced intervals. For loss, instead of using mean squared error (MSE) which is classically used for regression problems, we use log-cosh loss which helps in speeding up the convergence as compared to the work in Ref.  \cite{Raissi} and which is of the form $\frac{1}{N}\sum_{i=1}^N \ln(\cosh(\hat{y_i} - y_i))$. We further sketch out the details of this loss function and its advantages in Appendix \ref{appendix:experiments}. To optimize our model weights, we use Adam Optimizer with batch of size 100 and a fixed learning rate of $10^{-3}$. Given the simplicity of the payoff structure, for all our experiments, we use a 2-hidden layer architecture. For simplicity, we only construct TN layers that are symmetric in each input and each output dimension. In practice, we choose the first layer in our NN to be a dense layer with neurons that match the input shape of the second TN layer. That is, a DNN$(x,y)$ corresponds to a two-layer dense network with $x$ neurons in layer 1 and $y$ neurons in layer 2. On the other hand, a TNN$(x)$ corresponds to a two layer TNN architecture with the first being a dense layer with $x$ neurons and the second layer a TN layer with $x$ neurons. \newline

\noindent
We described in Section \ref{sec:tn} how TNN can compress dense layers, which was experimentally tested in Ref. \cite{TNN_NIPS}. However, this compression is only relevant if no DNN with the same parameter count can achieve the same accuracy and training performance. \newline

\begin{figure}[htp]
\centering
\includegraphics[scale=0.45]{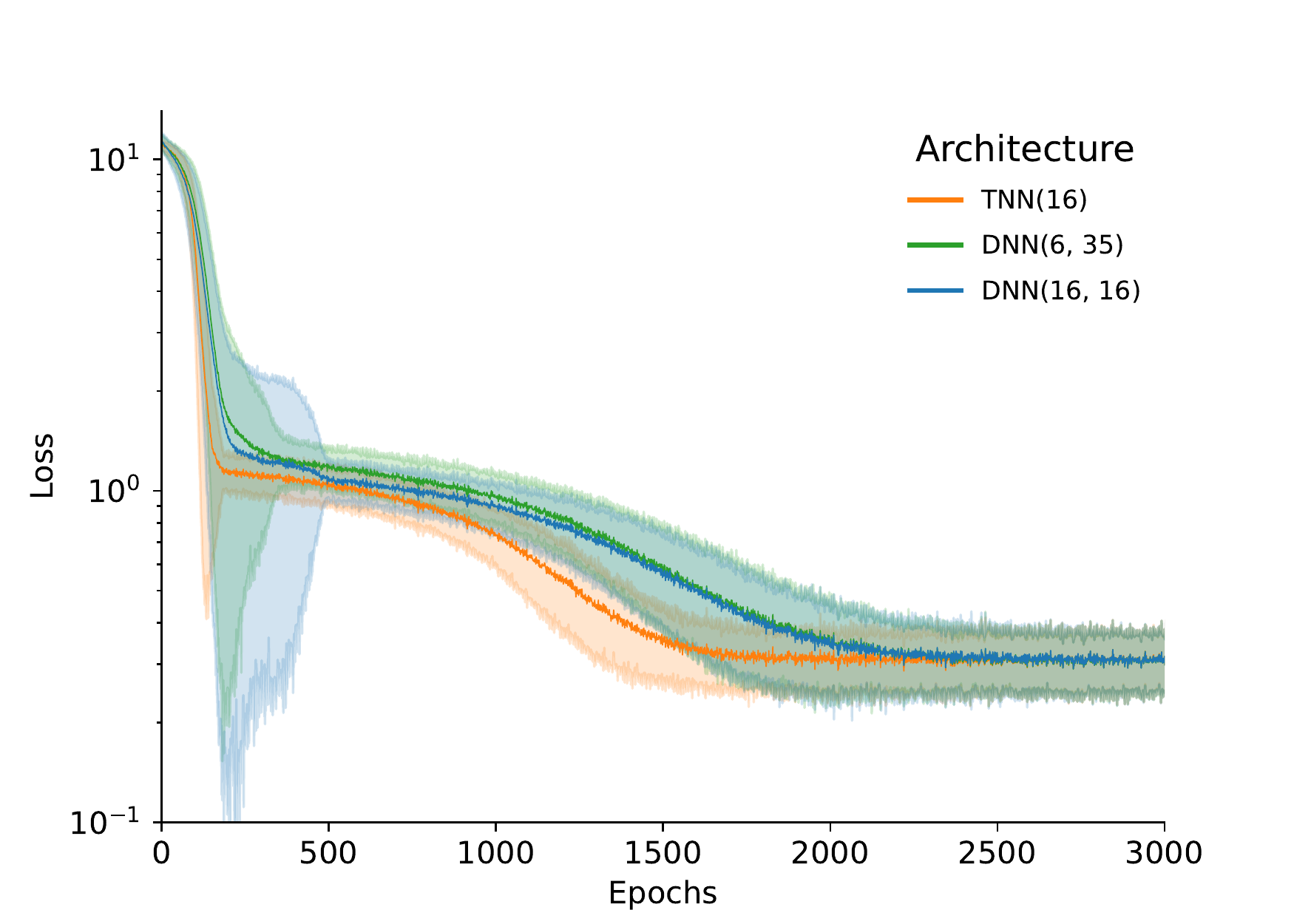} \hfill
\includegraphics[scale=0.45]{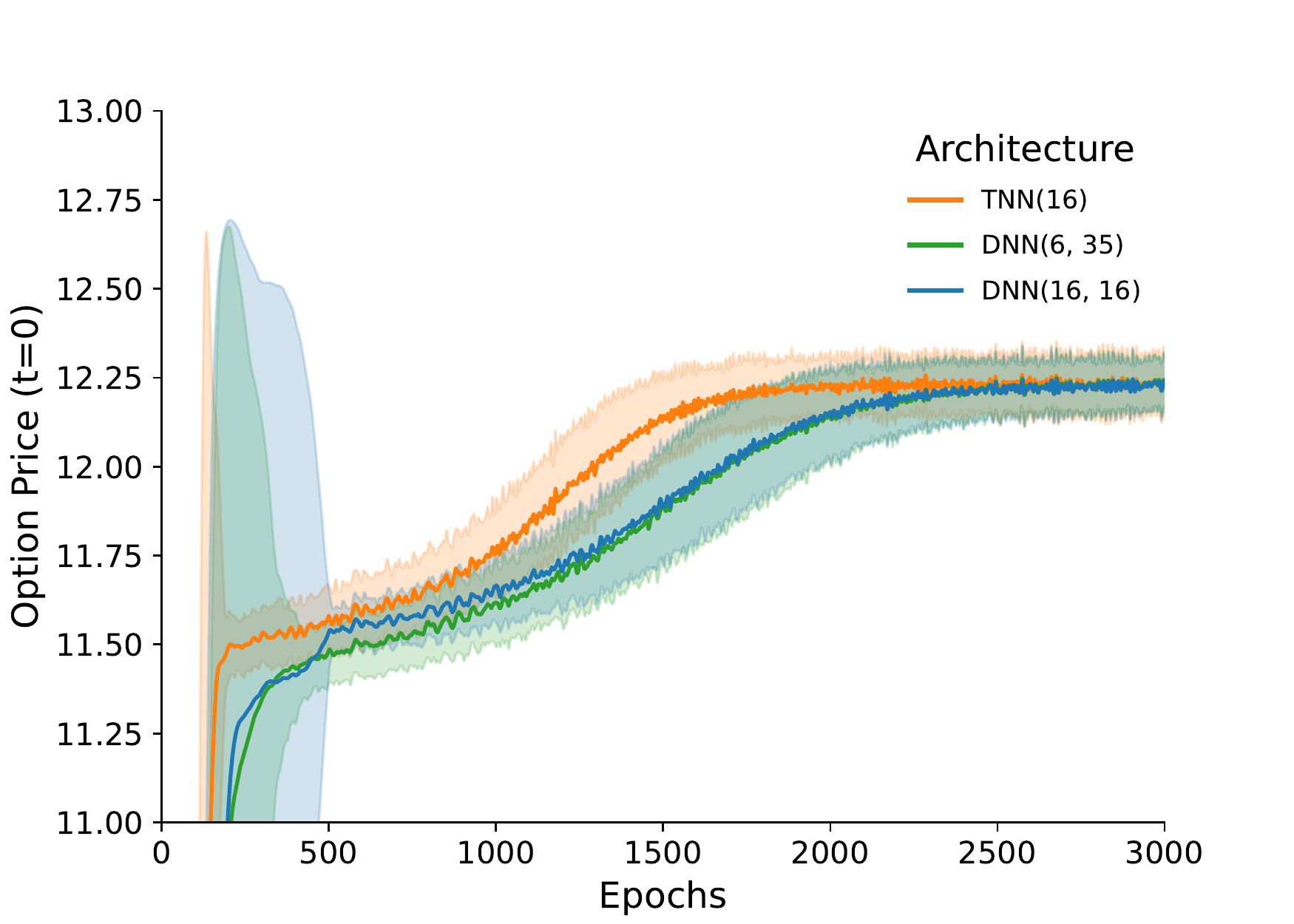}
\caption{(\textbf{left panel}) Training loss and (\textbf{right panel}) option price at $t_0$ over epochs of TNN(16) bond dimension 4 case (orange) and the corresponding best performing DNN with the same number of parameters (in this case, 353). The plots indicate resulting mean $\pm$ standard deviation from 100 runs.}
\label{Fig:Fig4}
\end{figure}

\noindent
In Fig.~\ref{Fig:Fig4}, we see the loss behavior and the option price at t = 0 for three different architectures, TNN(16), DNN(16,16) and DNN(6,35). Note that, in comparison with TNN(16), DNN(16,16) has the same number of neurons but more parameters. Whereas, DNN(6,35) has the same number of parameters but different number of neurons. All three architectures achieve the same accuracy level upon convergence. So, although TNN(16) is achieving the same accuracy as DNN(16,16) with fewer parameters, we find DNN(6,35) to be equally good in terms of accuracy and number of parameters. Hence, the number of parameters may not be used as a measure of compression without considering alternative DNN architectures with same parameter counts, which is a major drawback in the experiments performed in Ref. \cite{TNN_NIPS}.
\newline


\noindent
Moreover, we see in our experiments that the architectures differ in convergence speed. DNN(6,35) converges fastest among all the DNN architectures with the same parameter count as that of TNN(16). However, we observe that the TNN architecture converges even faster than DNN(6,35). It also converges faster than DNN(16,16). In summary, TNN not only allows for memory savings with respect to DNN for the same number of neurons, but also for faster convergence for the same number of parameters and neurons.\newline

\noindent
To get a better understanding of the advantages and behavior of TNN after taking various hyperparameters into account, we further implemented the following three experiments: (a) testing convergence speed for TNN with increasing bond dimension, (b) testing convergence speed for TNN with increasing number of neurons, and (c) finding the best matching DNN architecture (in terms of convergence speed and accuracy) and its parameter count as compared to a TNN.\newline


\begin{figure}[htp]
\centering
\includegraphics[scale=0.45]{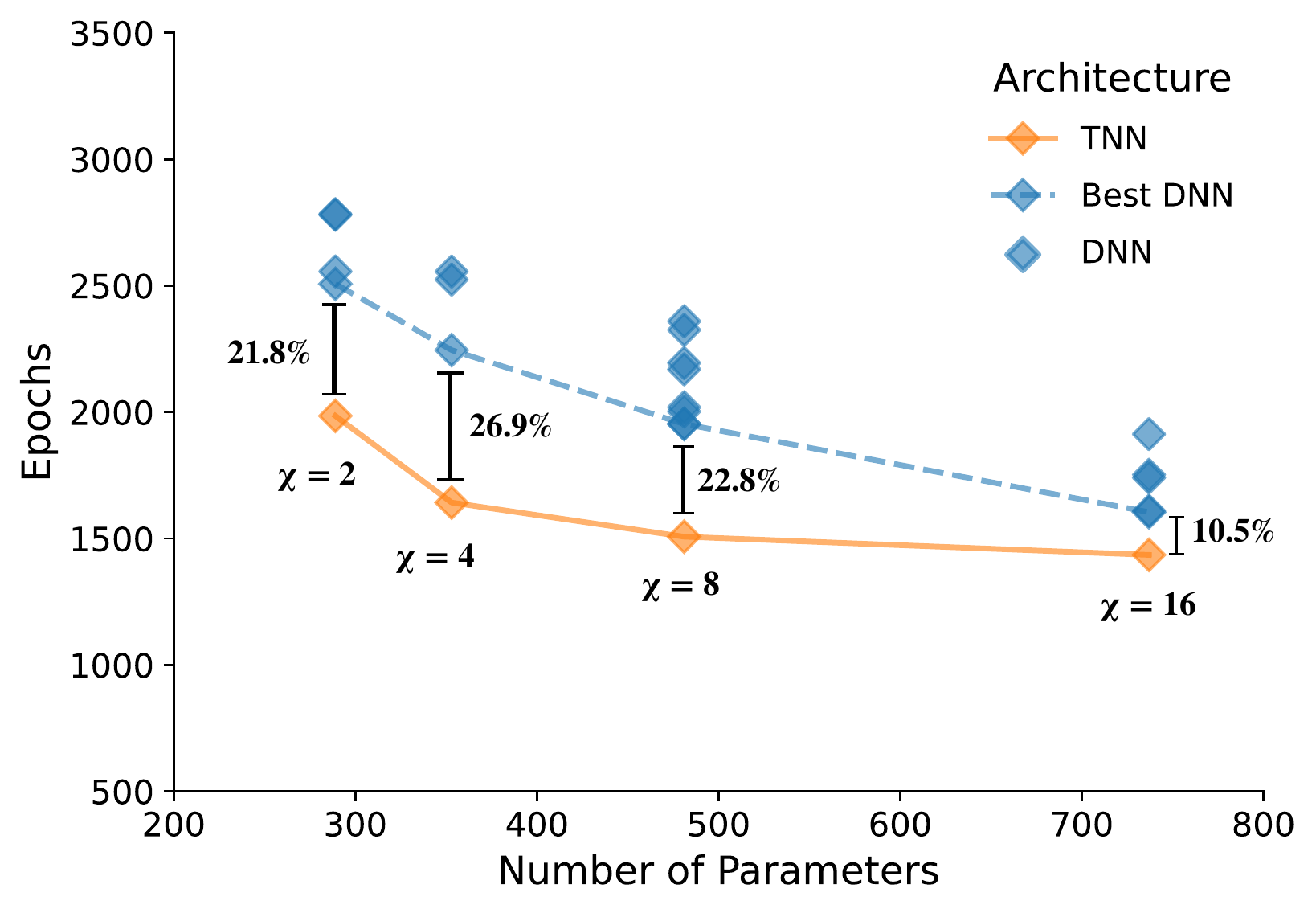}\hfill
\includegraphics[scale=0.45]{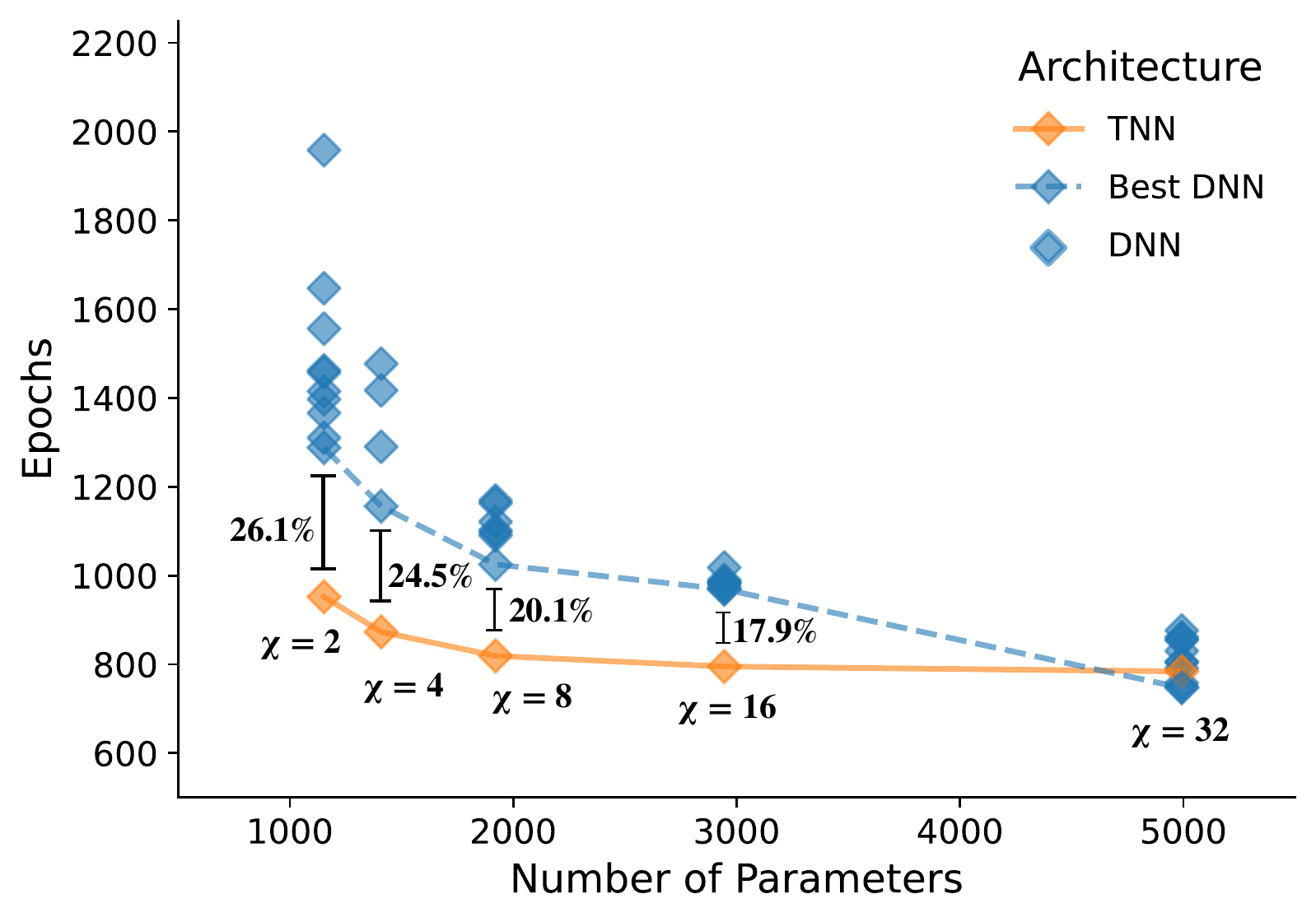}
\caption{Epoch for loss convergence for TNN(16) (\textbf{left panel}) and TNN(64) (\textbf{right panel}) over different bond dimensions. Each TNN is shown along with all DNNs of same number of parameters. The plots indicate results from 100 runs for the architectures reaching accuracy within 1\% of the true solution.}
\label{Fig:Fig5}
\end{figure}

\noindent
Let us discuss experiment (a), where we analyzed the effect of the bond dimension to the convergence behavior. We choose two TN architectures, TN(16) and TN(64) and vary their bond dimension $\chi$. For each $\chi$ we calculate the number of parameters in the network and construct all possible two-layer DNN with the same parameter count. We train the networks until convergence (convergence criteria is described in Appendix \ref{appendix:experiments}) and show the results in Fig.~\ref{Fig:Fig5}. We observe that almost all TNN architectures train faster than their DNN counterparts. The convergence gap is the largest for TNN(16) for $\chi = 4$ with 26.9\% fewer epochs to converge compared to the best performing DNN with the same parameter count. Furthermore, we see that with increasing bond dimension, the convergence gap diminishes, which is a signature that the corresponding MPO is approaching a dense layer and therefore the performance of a DNN as seen in our discussion in Section \ref{sec:tn}. An MPO with sufficiently large bond dimension can approximate any arbitrary matrix to an arbitrary degree. Hence, a TNN with a large bond dimension should exhibit a weight matrix in the forward pass that is very much comparable to an equally sized weight matrix of a DNN. \newline

\begin{figure}[htp]
\centering
\includegraphics[scale=0.45]{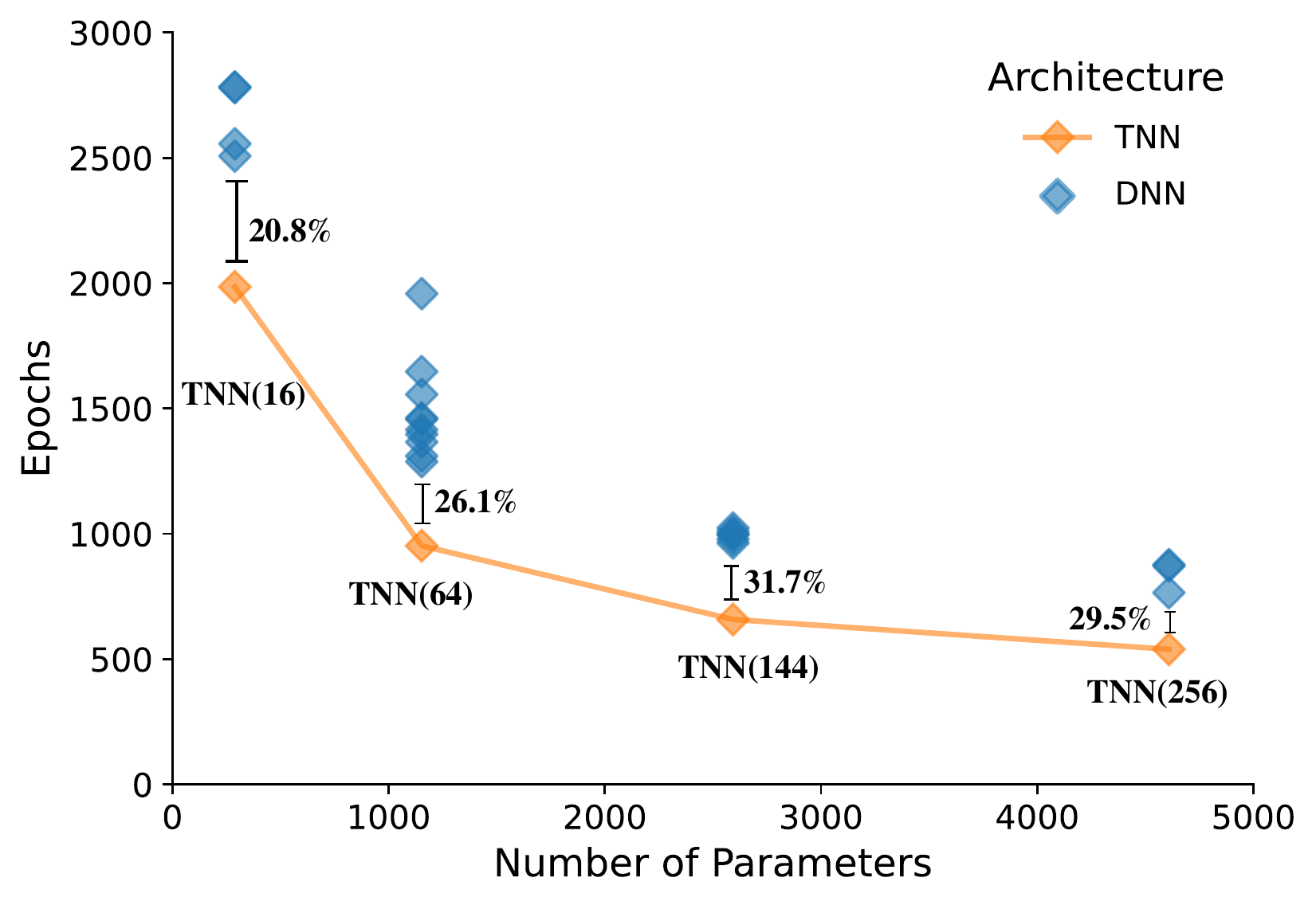}\hfill
\includegraphics[scale=0.45]{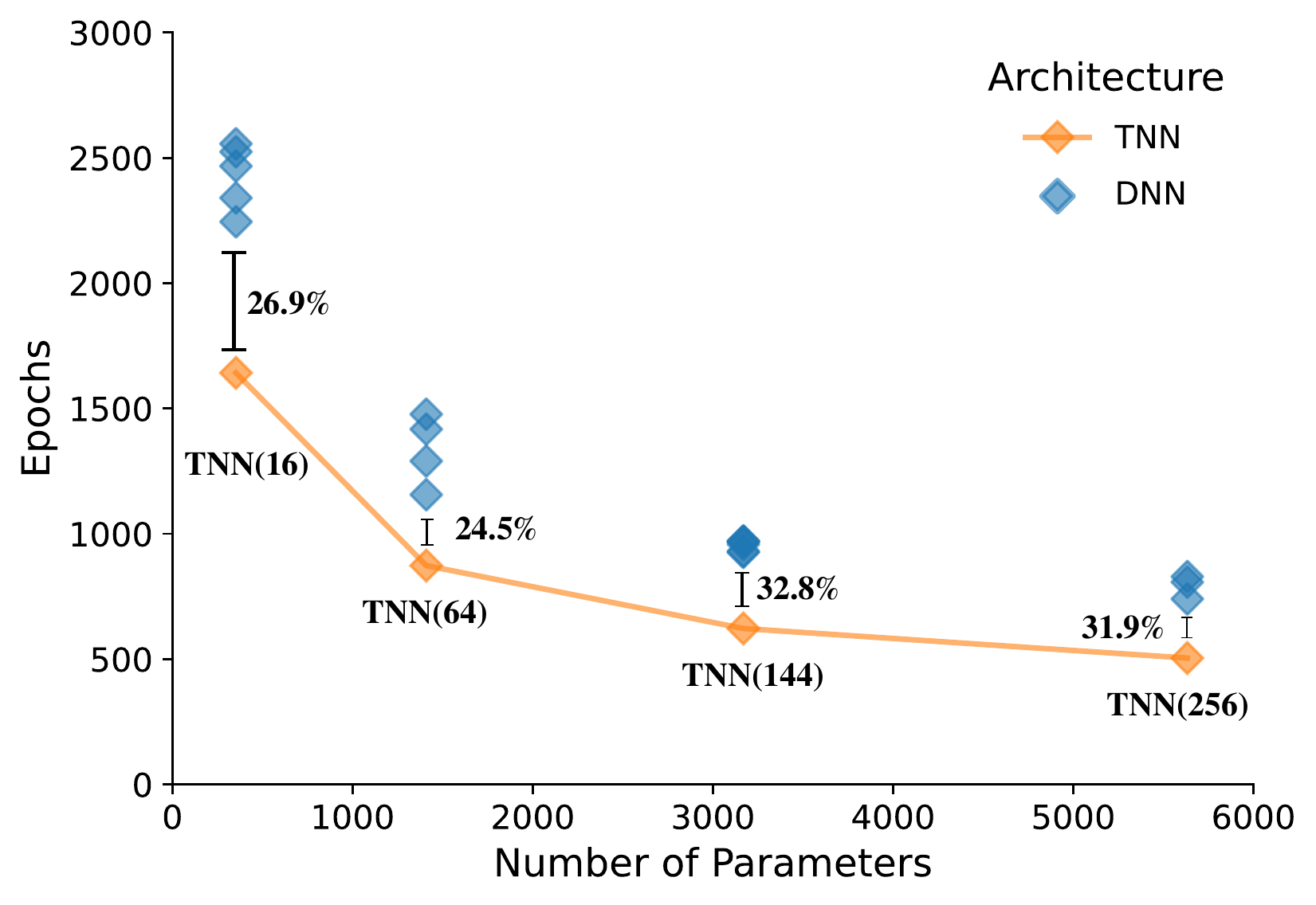}
\caption{Epoch for loss convergence for TNN of bond dimension 2 (\textbf{left panel}) and 4 (\textbf{right panel}) with varying number of layer width (TNN(16), TNN(64), TNN(144) and TNN(256)). Each TNN is shown along with all DNNs of same number of parameters. The plots indicate results from 100 runs for the architectures reaching accuracy within 1\% of the true solution.}
\label{Fig:Fig6}
\end{figure}

\noindent
Model complexity is another important factor when discussing the advantages of different architectures. We chose not to go deeper than two-layer networks due to the simplicity of the PDE at hand. However, we would like to know how TNN behaves if we construct wider networks. To this end, we increase the number of neurons in experiment (b). We choose $\chi = 2$ and $\chi = 4$ for the TNN we analyse and use the same convergence criteria as in experiment (a). The results are shown in Fig.~\ref{Fig:Fig6} for TNN(16), TNN(64), TNN(144) and TNN(256). In all cases, TNN is trained faster than its best DNN counterparts. We find the largest convergence gap of 32.8\% for TNN(144) with $\chi = 4$.  \newline

\begin{figure}
\centering
\includegraphics[height=6cm]{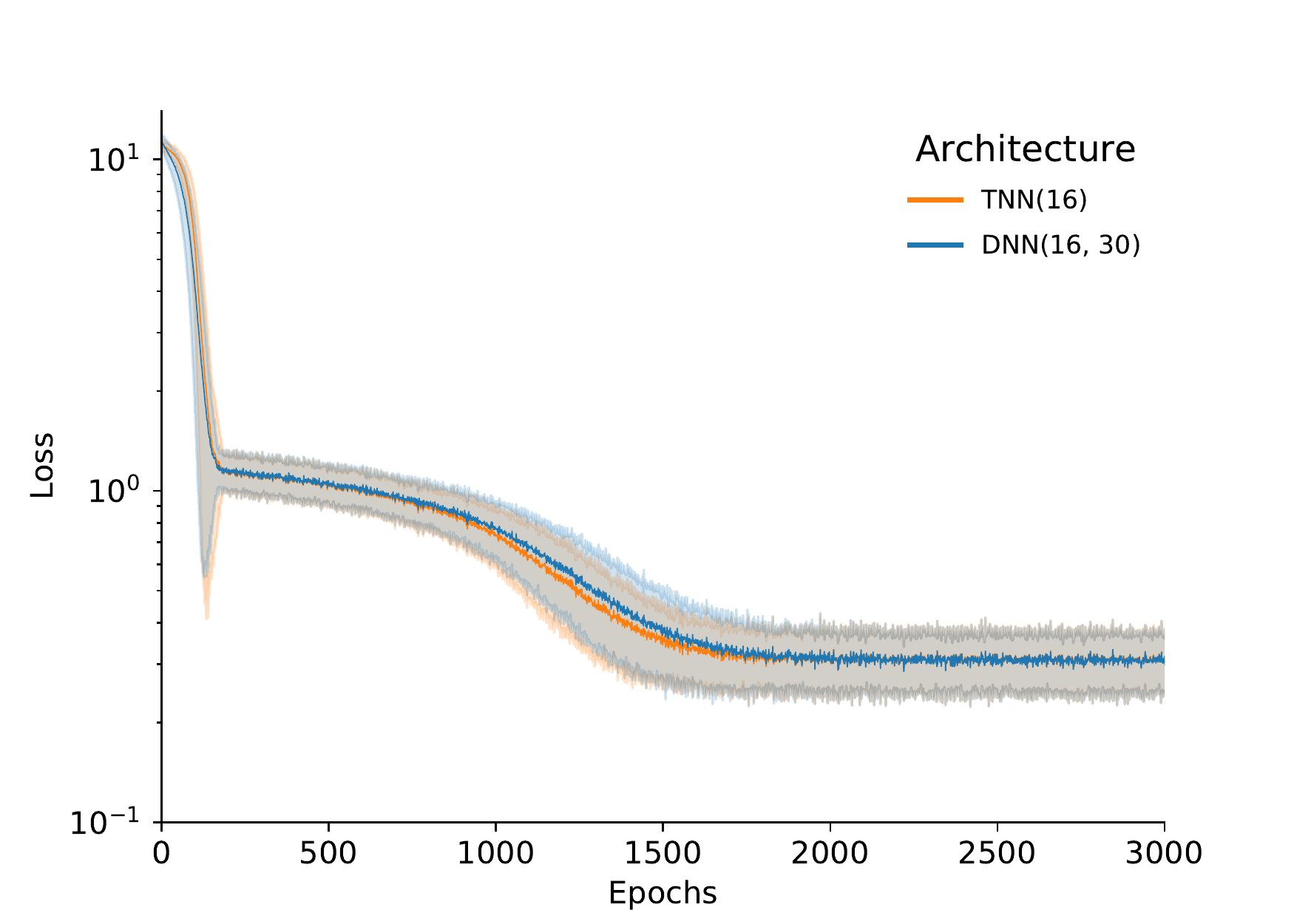}
\caption{Loss functions for TNN (orange) with bond dimension 4 and its best matching DNN (blue). The TNN has 353 parameters whereas the DNN has 1057 parameters. The plots indicate resulting mean $\pm$ standard deviation from 100 runs}
\label{fig:loss_matching_tnn_dnn}
\end{figure}

\noindent
Finally, in experiment (c), we analyse the parameter savings of the TNN for the example presented in Fig.~\ref{Fig:Fig4}. One way of quantifying this is to find the smallest DNN architecture that matches the loss signature of TNN, i.e., one that has the same training speed and accuracy. Fig.~\ref{fig:loss_matching_tnn_dnn} shows such an example, where we compare a DNN architecture with 1057 parameters to that of a TNN with 353 parameters. We conclude that \emph{with DNN we almost need three times the number of parameters to match the TNN's performance}. This parameter saving of 66\% is indeed a significant advantage for TNN. \newline

\noindent
To consolidate the TNN advantage observed in the plots from this section, we also tested TNN on the Hamilton-Jacobi-Bellman equation, which can be found in Appendix \ref{appendix:experiments}.   
\noindent

\section{Conclusions and Outlook}
\label{sec:conclude}
We have shown how we can leverage TNN to solve high-dimensional parabolic PDEs. Furthermore, we addressed some of the shortcomings in the existing literature when quantifying the advantages of TNN by analyzing parameter savings and convergence speed. Empirically, we demonstrated that TNN provides significant parameter savings as compared to a DNN while attaining the same accuracy. We further illustrated that TNN achieves speedup in training by comparing them against entire families of DNN architectures with similar parameter count. The methodology described in this paper can be used to improve training from a memory and speed perspective for a wide variety of problems in machine learning. Quantifying the complexity of a problem and adapting the methodology presented to problems where this approach can provide a significant edge, can be an interesting avenue for future work.

\bigskip 
\noindent
{\bf Acknowledgements -} We acknowledge regular fruitful discussions with the technical teams both at Cr\'edit Agricole and Multiverse Computing. 
\newpage
\bibliographystyle{abbrv}
\bibliography{references}

\begin{thebibliography}{10}

\bibitem{Dataaug}
A.~M. Aboussalah, M.-J. Kwon, R.~G. Patel, C.~Chi, and C.-G. Lee.
\newblock Don't overfit the history -- recursive time series data augmentation.
\newblock {\em arXiv preprint arXiv:2207.02891}, 2022.

\bibitem{Antonelli}
F.~Antonelli.
\newblock Backward-forward stochastic differential equations.
\newblock {\em Annals of Applied Probability}, pages 777--793, 1993.

\bibitem{bhatia2019matrix}
A.~Bhatia, M.~K. Saggi, A.~Kumar, and S.~Jain.
\newblock Matrix product state--based quantum classifier.
\newblock {\em Neural computation}, 31(7):1499--1517, 2019.

\bibitem{bradley2020modeling}
T.-D. Bradley, E.~M. Stoudenmire, and J.~Terilla.
\newblock Modeling sequences with quantum states: A look under the hood.
\newblock {\em Machine Learning: Science and Technology}, 2020.

\bibitem{PhysRevB.99.155131}
S.~Cheng, L.~Wang, T.~Xiang, and P.~Zhang.
\newblock Tree tensor networks for generative modeling.
\newblock {\em Physical Review B}, 99:155131, 2019.

\bibitem{Cheridito}
P.~Cheridito, H.~M. Soner, T.~Nizar, and N.~Victoir.
\newblock Second-order backward stochastic differential equations and fully
  nonlinear parabolic \uppercase{PDE}s.
\newblock {\em Communications on Pure and Applied Mathematics},
  60(7):1081--1110, nov 2006.

\bibitem{Beck_2019}
B.~Christian, E.~Weinan, and J.~Arnulf.
\newblock Machine learning approximation algorithms for high-dimensional fully
  nonlinear partial differential equations and second-order backward stochastic
  differential equations.
\newblock {\em Journal of Nonlinear Science}, 29(4):1563--1619, jan 2019.

\bibitem{quasi-FBPDE}
F.~Delarue and S.~Menozzi.
\newblock A forward-backward stochastic algorithm for quasi-linear
  \uppercase{PDE}s.
\newblock {\em The Annals of Applied Probability}, pages 140--184, 2006.

\bibitem{NIPS2013_7fec306d}
M.~Denil, B.~Shakibi, L.~Dinh, M.~A. Ranzato, and N.~de~Freitas.
\newblock Predicting parameters in deep learning.
\newblock {\em Advances in Neural Information Processing Systems},
  26:2148--2156, 2013.

\bibitem{Stoudenmire_2018}
S.~Edwin.
\newblock Learning relevant features of data with multi-scale tensor networks.
\newblock {\em Quantum Science and Technology}, 3(3):034003, 2018.

\bibitem{NIPS2016_6211}
S.~Edwin and D.~J. Schwab.
\newblock Supervised learning with tensor networks.
\newblock {\em Advances in Neural Information Processing Systems 29}, pages
  4799--4807, 2016.

\bibitem{efthymiou2019tensornetwork}
S.~Efthymiou, J.~Hidary, and S.~Leichenauer.
\newblock Tensor network for machine learning.
\newblock {\em arXiv preprint arXiv:1906.06329}, 2019.

\bibitem{glasser2018supervised}
I.~Glasser, N.~Pancotti, and J.~I. Cirac.
\newblock Supervised learning with generalized tensor networks.
\newblock {\em arXiv preprint arXiv:1806.05964}, 2018.

\bibitem{9058650}
I.~Glasser, N.~Pancotti, and J.~I. Cirac.
\newblock From probabilistic graphical models to generalized tensor networks
  for supervised learning.
\newblock {\em IEEE Access}, 8:68169--68182, 2020.

\bibitem{PhysRevX.8.031012}
Z.-Y. Han, J.~Wang, H.~Fan, L.~Wang, and P.~Zhang.
\newblock Unsupervised generative modeling using matrix product states.
\newblock {\em Physical Review X}, 8:031012, 2018.

\bibitem{Han_2018}
H.~Jiequn, J.~Arnulf, and E.~Weinan.
\newblock Solving high-dimensional partial differential equations using deep
  learning.
\newblock {\em Proceedings of the National Academy of Sciences},
  115(34):8505--8510, aug 2018.

\bibitem{Liu_2019}
D.~Liu, S.-J. Ran, P.~Wittek, C.~Peng, R.~B. Garc{\'{\i}}a, G.~Su, and
  M.~Lewenstein.
\newblock Machine learning by unitary tensor network of hierarchical tree
  structure.
\newblock {\em New Journal of Physics}, 21(7):073059, 2019.

\bibitem{4-stepPDE}
J.~Ma, P.~Protter, and J.~Yong.
\newblock Solving forward-backward stochastic dif- ferential equations
  explicitly: a four step scheme.
\newblock {\em Probability theory and related fields}, 98:339--359, 1994.

\bibitem{TNN_NIPS}
A.~Novikov, D.~Podoprikhin, A.~Osokin, and D.~P. Vetrov.
\newblock Tensorizing neural networks.
\newblock {\em Advances in Neural Information Processing Systems}, 28, 2015.

\bibitem{RomanTN}
R.~Orús.
\newblock A practical introduction to tensor networks: Matrix product states
  and projected entangled pair states.
\newblock {\em Annals of Physics}, 349:117--158, 2014.

\bibitem{tnn_osedelets}
I.~V. Oseledets.
\newblock Tensor-train decomposition.
\newblock {\em SIAM Journal on Scientific Computing}, 33(5):2295--2317, 2011.

\bibitem{quasi-PDE}
E.~Pardoux and S.~Tang.
\newblock Forward-backward stochastic differential equations and quasilinear
  parabolic \uppercase{PDE}s.
\newblock {\em Probability theory and related fields}, 114:123--150, 1999.

\bibitem{Raissi}
M.~Raissi.
\newblock Forward-backward stochastic neural networks: Deep learning of
  high-dimensional partial differential equations.
\newblock {\em arXiv preprint arXiv:1804.07010v1}, 2018.

\bibitem{Raissi-part1}
M.~Raissi, P.~Perdikaris, and G.~E. Karniadakis.
\newblock Physics informed deep learning (part i): Data-driven solutions of
  nonlinear partial differential equations.
\newblock {\em arXiv preprint arXiv:1711.10561}, 2017.

\bibitem{Raissi-part2}
M.~Raissi, P.~Perdikaris, and G.~E. Karniadakis.
\newblock Physics informed deep learning (part ii): Data-driven discovery of
  nonlinear partial differential equations.
\newblock {\em arXiv preprint arXiv:1711.10566}, 2017.

\bibitem{low_rank}
T.~N. Sainath, B.~Kingsbury, V.~Sindhwani, E.~Arisoy, and B.~Ramabhadran.
\newblock Low-rank matrix factorization for deep neural network training with
  high-dimensional output targets.
\newblock {\em International Conference on Acoustics, Speech and Signal
  Processing}, pages 6655--6659, 2013.

\bibitem{tn_memory}
K.~Simonyan and A.~Zisserman.
\newblock Very deep convolutional networks for large-scale image recognition.
\newblock {\em International Conference on Learning Representations ({ICLR})},
  2015.

\bibitem{PhysRevB.101.075135}
Z.-Z. Sun, C.~Peng, D.~Liu, S.-J. Ran, and G.~Su.
\newblock Generative tensor network classification model for supervised machine
  learning.
\newblock {\em Physical Review B}, 101:075135, 2020.

\bibitem{tebd}
G.~Vidal.
\newblock Efficient simulation of one-dimensional quantum many-body systems.
\newblock {\em Phys. Rev. Lett.}, 93:040502, Jul 2004.

\bibitem{E_2017}
E.~Weinan, H.~Jiequn, and J.~Arnulf.
\newblock Deep learning-based numerical methods for high-dimensional parabolic
  partial differential equations and backward stochastic differential
  equations.
\newblock {\em Communications in Mathematics and Statistics}, 5(4):349--380,
  nov 2017.

\bibitem{dmrg}
S.~R. White.
\newblock Density matrix formulation for quantum renormalization groups.
\newblock {\em Phys. Rev. Lett.}, 69:2863--2866, Nov 1992.

\bibitem{xue13_interspeech}
J.~Xue, J.~Li, and Y.~Gong.
\newblock {Restructuring of deep neural network acoustic models with singular
  value decomposition}.
\newblock {\em Interspeech, 2013}, pages 2365--2369, 2013.

\end{thebibliography}

\newpage

\appendix 
\noindent 
\Large \textbf{Appendix}
\normalsize

\section{Feynman-Kac Representation}
\vspace{-0.2cm}
\label{F-K}
\noindent
Let $\mu:\RR_+\times \RR^d\to \RR^d$, $\sigma:\RR_+\times \RR^d\to \RR^{d\times m}$, $V:\RR_+\times \RR^d\to \RR$ and $f:\RR^d\to \RR$ be sufficiently regular functions, fix $T>0$ and suppose $u\in C^{1,2}([0,T]\times \RR^d\to \RR)$ is the unique solution to the parabolic differential equation
\begin{equation}\label{eqn: parabolic_PDE}
\begin{aligned}
\frac{\partial u}{\partial t}+Au&=Vu\\
u(T,x)&=f(x)
\end{aligned}
\end{equation}
with generator defined by
\begin{equation}\label{eqn:generator}
Ag(t,x)=\sum_{i=1}^d \mu_i(t,x)\frac{\partial g}{\partial x_i}(t,x)+\frac{1}{2}\sum_{i,j=1}^d \left(\sigma\sigma^T \right)_{ij}(t,x)\frac{\partial^2g}{\partial x_i\partial x_j}(t,x) 
\end{equation}
for $g\in C^{1,2}([0,T]\times \RR^d\to \RR)$. If $Y$ is the $\RR^d$-valued stochastic process satisfying the stochastic differential equation
$$d Y_t=\mu(t,Y_t)d t+\sigma(t,Y_t)d B_t$$
for some $m$-dimensional Brownian motion $B$ defined on a probability space $(\O,\F,\P)$, then $u$ admits the stochastic representation
$$u(t,x)=\E^\P\left[e^{-\int_t^T V(s,X_s)d s}f(Y_T)\given Y_t=x \right]$$
for all $(t,x)\in [0,T]\times \RR^d$.
\vspace{-0.2cm}

\section{Further experiments}
\label{appendix:experiments}
\subsection{Hamilton-Jacobi-Bellman Equation} 
\vspace{-0.2cm}
Dynamic Decision Making (DDM) lies at the heart of a lot of industrial problems spanning from inventory control and robotics to finance. Integral to DDM is the difference equation called Bellman equation which is an optimality condition associated with dynamic programming, an approach which has been fundamental in solving DDM problems recursively. This Bellman equation in continuous form leads us to the Hamilton-Jacobi-Bellman (HJB) PDE which is an indispensable tool in the area of stochastic control problems. Here, we consider a 100-dimensional HJB equation as shown in Ref. \cite{Raissi} which looks as follows: 
\begin{equation}
    u_t = -{\rm Tr}\left( \mathcal{D}^2 u \right) + \|\mathcal{D}u\|^2,
\end{equation}

\noindent
with terminal condition $u(T,x) = \ln\left(0.5(1+\|x\|^2)\right)$. The explicit solution for this equation is
\begin{equation}
    u(t,x) = -\ln\left(\mathbb{E}\left[{\rm exp}(-g(x+\sqrt{2}W_{T-t}))\right]\right), 
\end{equation}
which can be used to test the accuracy of the proposed algorithm.  However, due to the presence of the expectation operator in the explicit solution, we can only approximate the exact solution. To do so, we use $10^5$ samples to approximate the exact solution as done in Ref.  \cite{Raissi}. Using Feynman-Kac as shown in Section \ref{NN-Math}, this can be recasted into a system of forward-backward stochastic differential equations as follows:
\vspace{-0.4cm}

\begin{align}
\begin{split}
dX_t &= \sigma dW_t, ~~~ t \in [0,T] \\
X_0 &= \epsilon \\
dY_t &= \|Z_t\|^2 dt + \sigma Z_t ' dW_t, ~~~ t \in[0,T)\\
Y_T &= \ln\left(0.5\left(1+\|X_T\|^2\right)\right),\\
\end{split}
\end{align}

\begin{figure}[htp]
\centering
\includegraphics[scale=0.4]{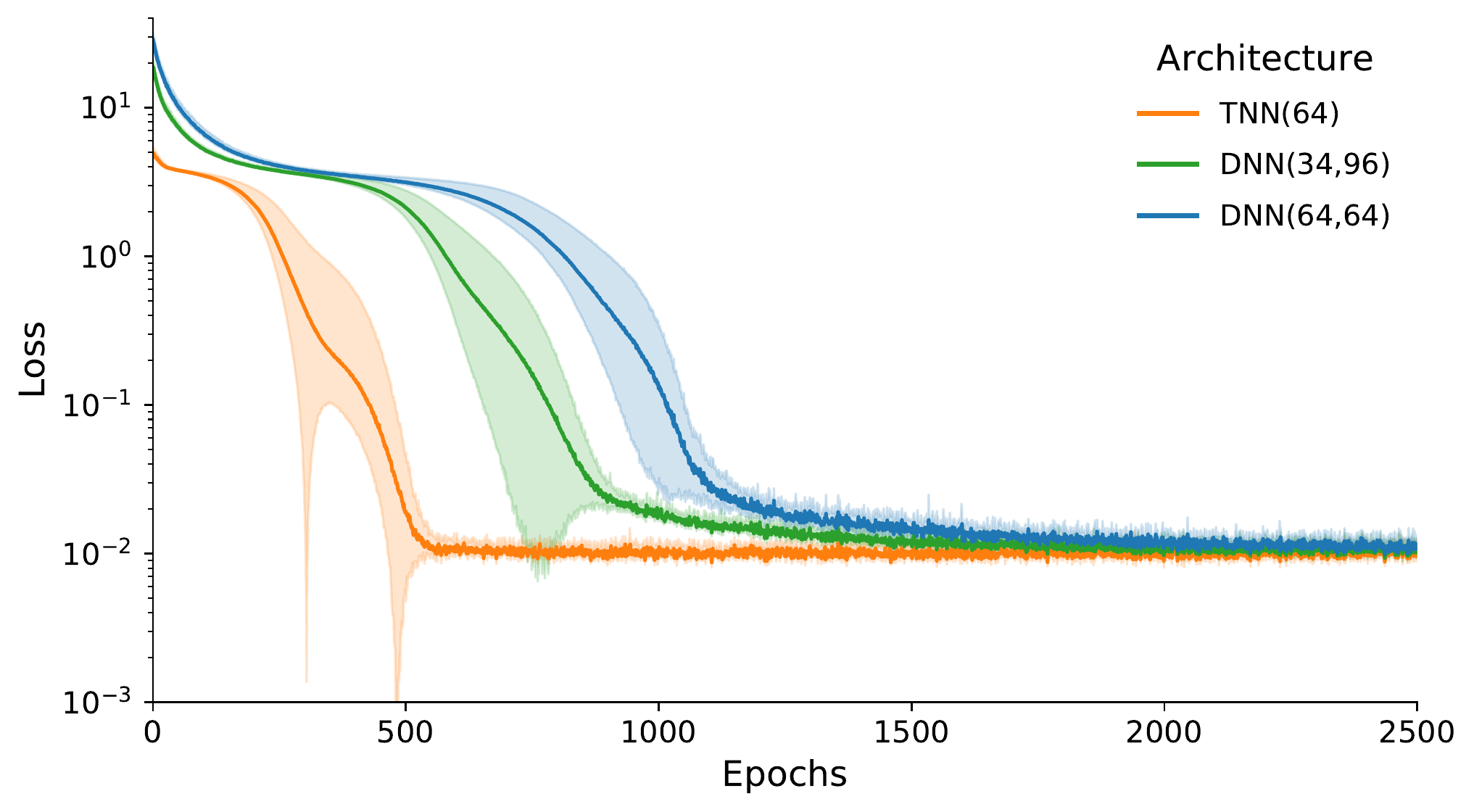} \hfill
\includegraphics[scale=0.4]{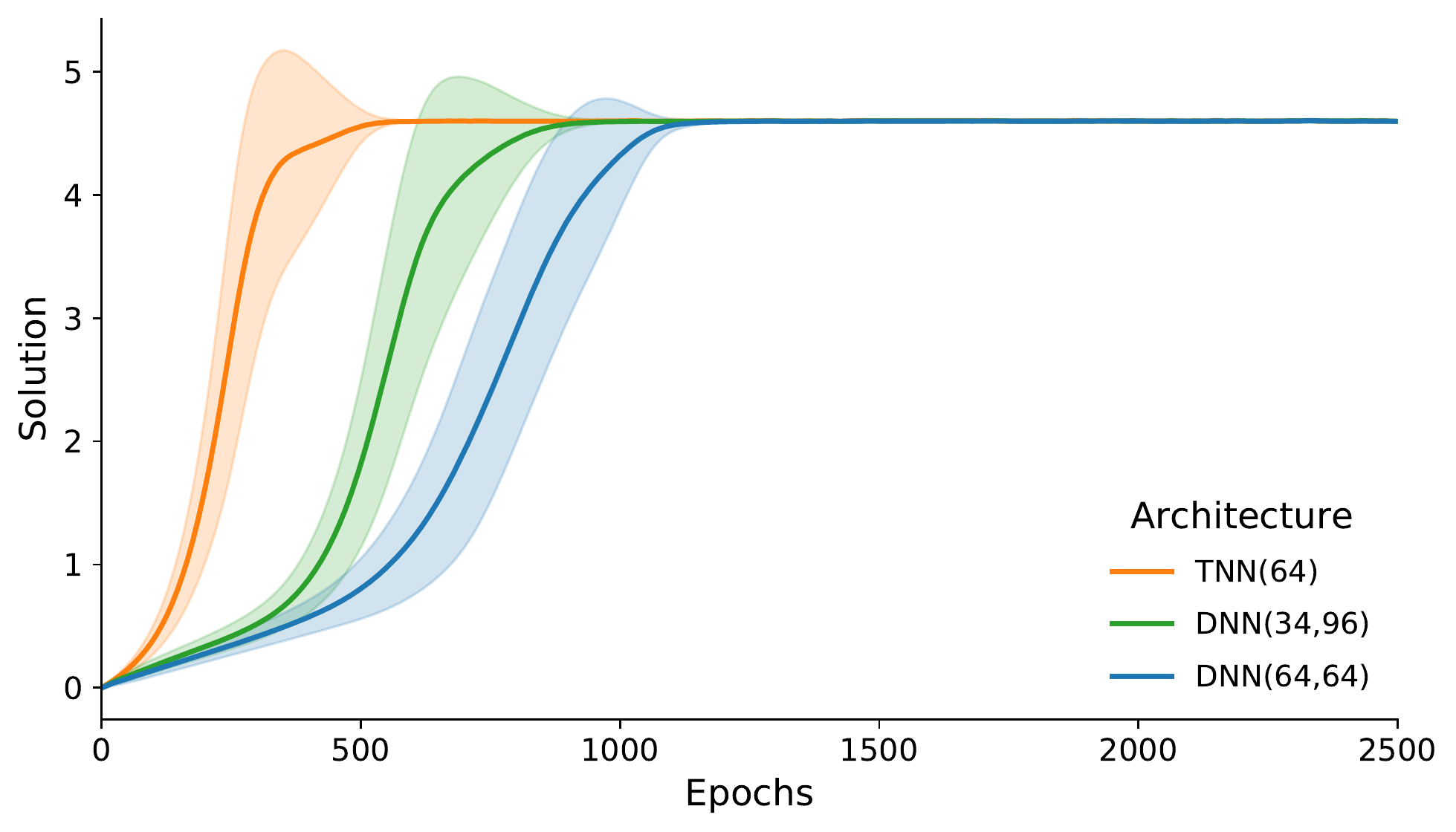}
\caption{(\textbf{left panel}) Training loss and (\textbf{right panel}) Solution for 100-d HJB equation at $t_0$ over epochs for TNN(64) with bond dimension 2 (orange), the corresponding best DNN with equivalent parameters (blue) and the DNN with equivalent neurons (green).}
\label{fig:loss_hjb}
\end{figure}

\noindent 
where $\epsilon = (0,0,0, \dots, 0) \in \mathbb{R}^{100}$, $T=1$, $\sigma=\sqrt{2}$. Furthermore, we partition the time domain $[0, T]$ into $N = 50$ equally spaced intervals. For loss, here too, instead of using mean squared error (MSE) which is classically used for regression problems, we use log-cosh loss which helps in speeding up the convergence as compared to the work in Ref. \cite{Raissi} and which is of the form $\frac{1}{N}\sum_{i=1}^N \ln(\cosh(\hat{y_i} - y_i))$. To optimize our model weights, we use Adam Optimizer with batch of size 100 and a fixed learning rate of $10^{-3}$. \newline

\noindent
\noindent
As shown in Fig.~\ref{fig:loss_hjb}, a TNN of 64 units and bond dimension 2 clearly outperforms all DNNs with the same parameter count in terms of convergence speed, thereby showing TNN’s clear advantage (in this plot, we plot DNN(34,96) which has the best performance among the DNNs with the same parameter count as TNN). Here, the advantage is even more observable, with TNN converging in 66\% fewer epochs (which is 63\% savings in time taken to converge). In the plot, we also show the comparison of TNN with a DNN with same number of neurons, i.e., DNN(64,64). Despite the DNN having same number of neurons and rather more parameters, TNN clearly outperforms it.

\subsection{Same Initialization Magnitude}
\begin{figure}[htp]
\centering
\includegraphics[scale=0.45]{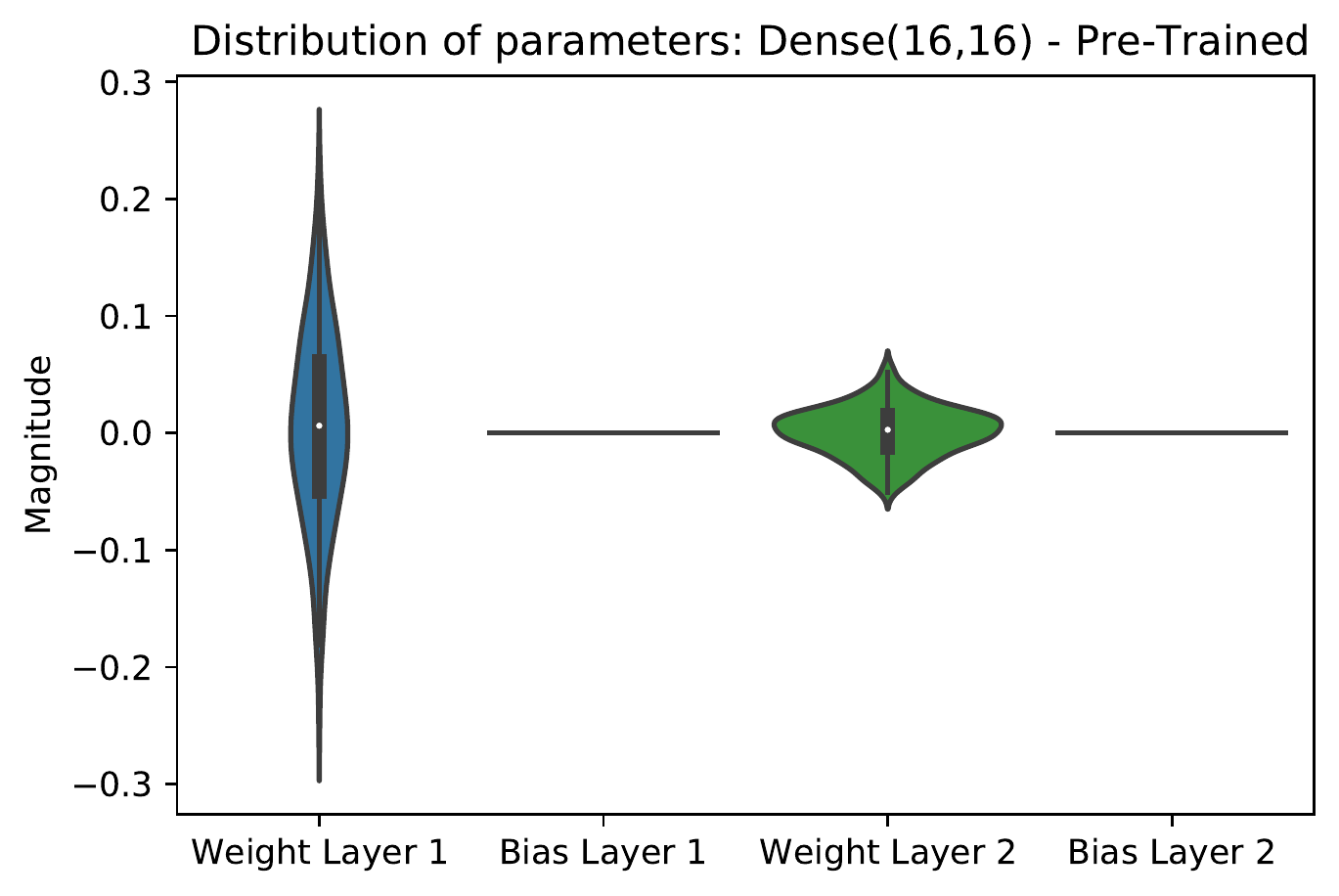}\quad
\includegraphics[scale=0.45]{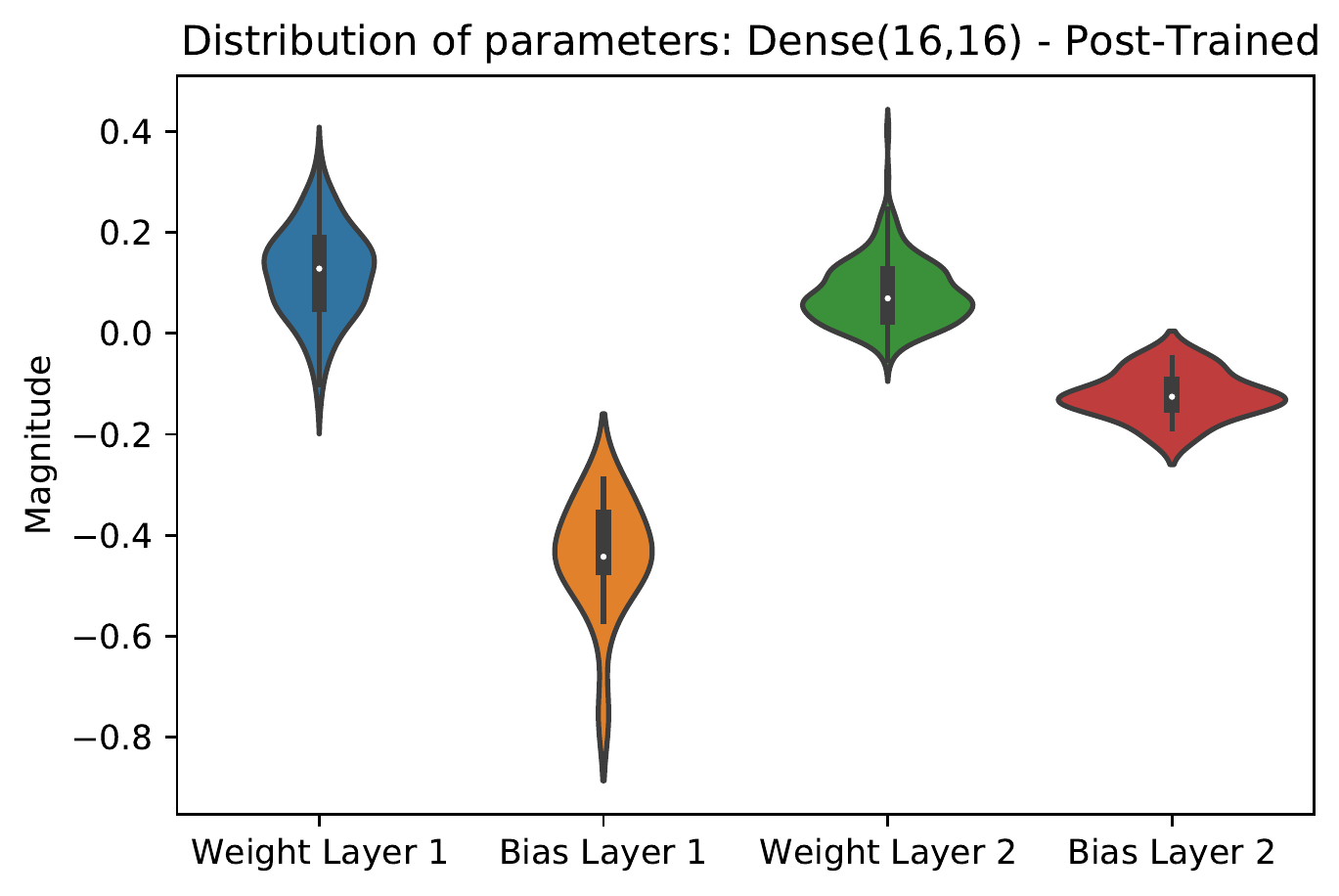} 
\caption{Pre-trained weights (left) and Post-trained weights (right) for a dense architecture with (16,16) neurons in the two hidden layers}
\label{fig:figure_dense_weights}
\end{figure}
\begin{figure}[htp]
\centering
\includegraphics[scale=0.45]{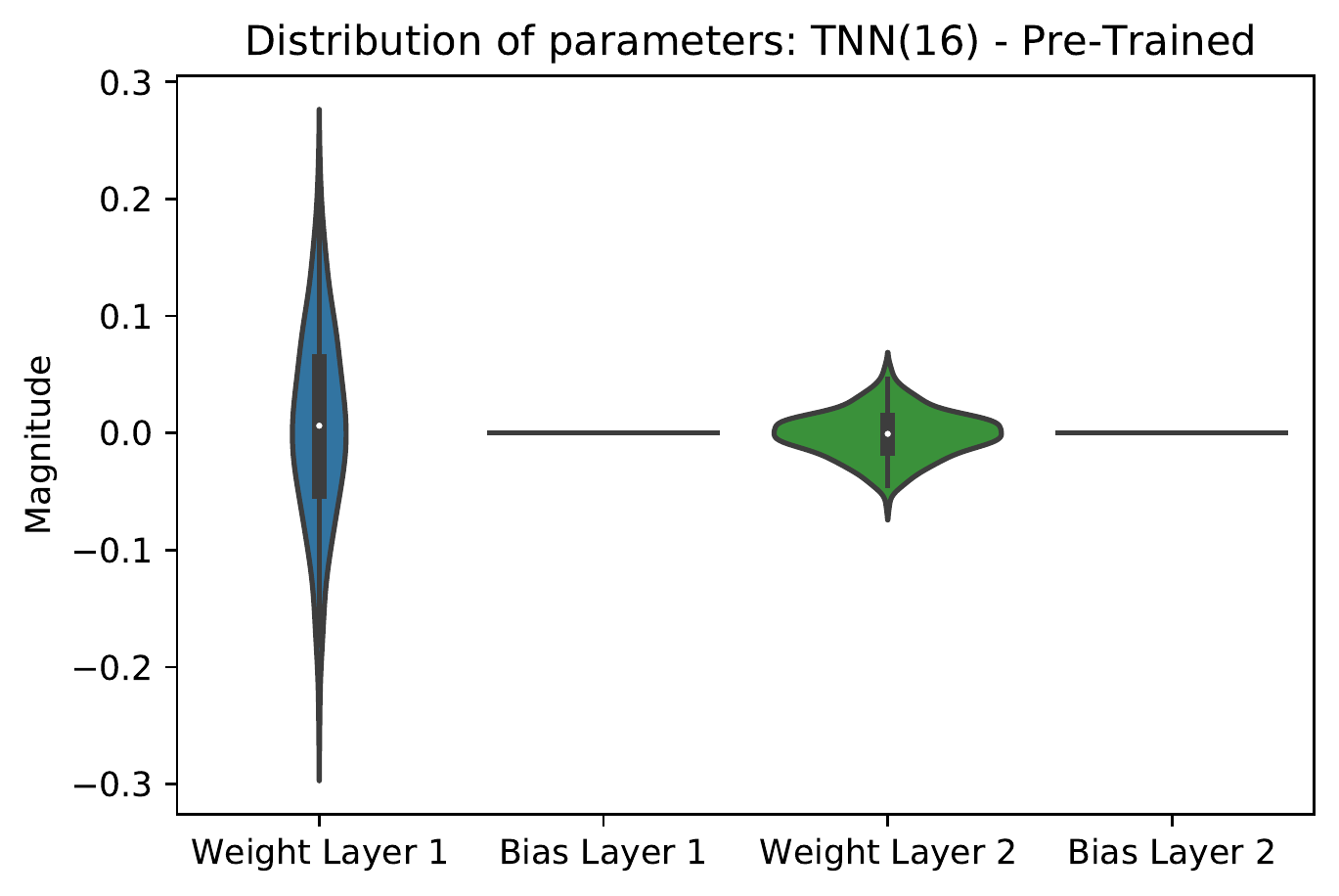}\quad
\includegraphics[scale=0.45]{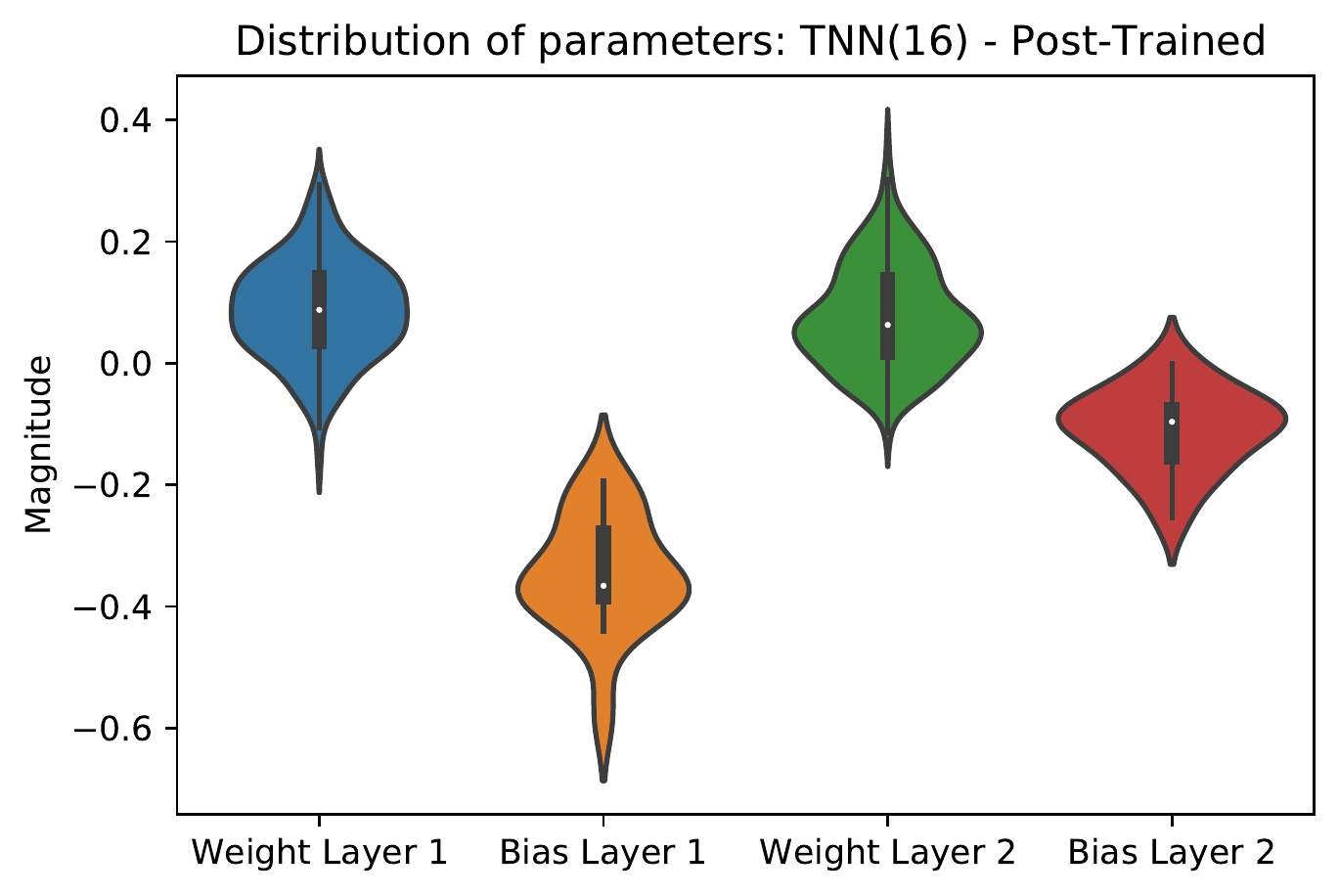} 
\caption{Pre-trained weights (left) and Post-trained weights (right) for a TNN architecture with bond dimension 8 and 2 tensors. This architecture has same number of parameters as the one in Fig.~\ref{fig:figure_dense_weights}}
\label{fig:figure_TN_weights}
\end{figure}
In Section \ref{sec:tn}, we saw that, for a TNN layer, we first initialize a set of tensors and then contract them to a shape that is equivalent to that of a dense layer. In doing so, we realize that if the tensors are initialized in a similar way as the weight matrix of a DNN layer, the magnitude of initialized weights is different when comparing the contracted tensor of a TNN layer with a weight matrix from a DNN layer with same size. This can often lead to different training trajectories. However, to ensure that the resulting advantage stems from the underlying structure of TNN and not magnitude of initialized weights, we also initialize them in a way to ensure similar magnitude. Here, we start by setting up weights in a way that the distribution of the pre-trained weights of a DNN layer and that of an equivalent contracted TNN layer with same number of parameters are similar as shown in Figs.~\ref{fig:figure_dense_weights} and \ref{fig:figure_TN_weights}. Despite having the same pre-trained weights, we still observe the TNN advantage as Fig.~\ref{Fig:Fig4}, which can also be attributed to different post-trained weight distribution. \newline 



\subsection{Hybrid Loss Function}

\begin{figure}[htp]
    \centerline{\includegraphics[scale=0.45]{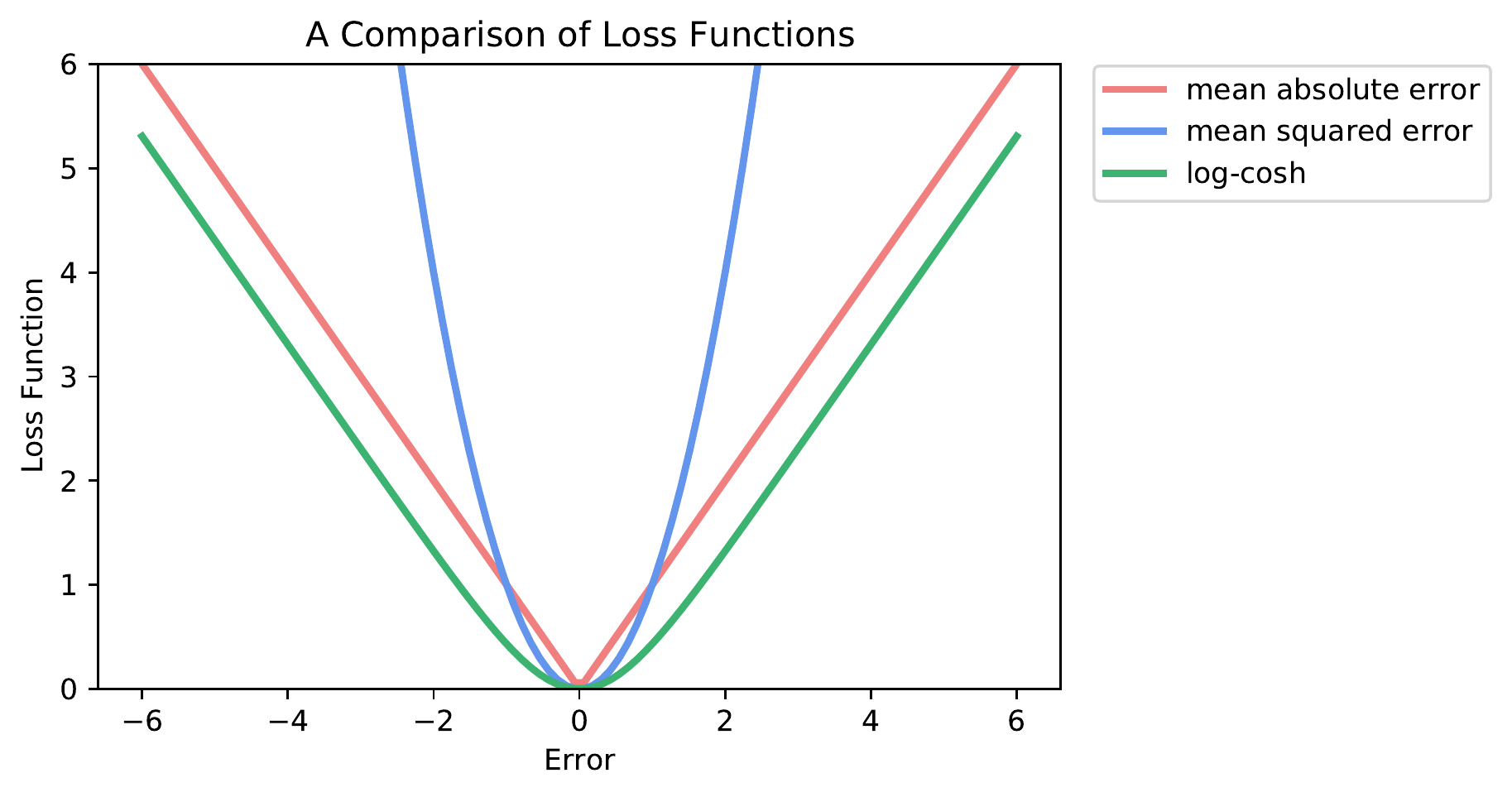}}
    \caption{Comparison of Mean Squared Error (MSE), Mean Absolute Error (MAE) and Hybrid loss for regression-type problems}
    \label{fig:hybrid_loss}
\end{figure}
For our experiments in this paper, we use a hybrid loss function over MSE due to its empirically enhanced speed of convergence as observed in our experiments. We use log-cosh loss of the form $\frac{1}{N}\sum_{i=1}^N \ln(\cosh(\hat{y_i} - y_i))$. The function $\ln(\cosh(x))$ for small $x$ turns out to be $\frac{x^2}{2}$, whereas for large $x$ this becomes $\lvert x \rvert - \ln(2)$. Comparing it to MSE, it looks as in Fig.~\ref{fig:hybrid_loss}. However, due to empirical results, we only apply this to the terminal conditions. As a result, our loss function changes from

\begin{multline}
\underset{\theta}{\rm min}\Bigg( \sum_{m=1}^M \sum_{n=0}^{N-1} \lvert Y_{n+1}^m (\theta) - Y_{n}^m (\theta) - \varphi(t_n, X_n ^m, Y_n ^m(\theta),Z_n ^m(\theta)) \Delta t_n  \\ 
- (Z_n ^m (\theta))^T \sigma(t_n, X_n ^m, Y_n ^m(\theta)) \Delta W_n ^m \rvert ^2+ \sum_{m=1}^M (Y_N ^m(\theta) - g(X_N ^m))^2\Bigg)
\end{multline}

to

\begin{multline}
\underset{\theta}{\rm min} \Bigg(\sum_{m=1}^M \sum_{n=0}^{N-1} \lvert Y_{n+1}^m (\theta) - Y_{n}^m (\theta) - \varphi(t_n, X_n ^m, Y_n ^m(\theta),Z_n ^m(\theta)) \Delta t_n  \\ 
- (Z_n ^m (\theta))^T \sigma(t_n, X_n ^m, Y_n ^m(\theta)) \Delta W_n ^m \rvert ^2+ \frac{1}{M}\sum_{m=1}^M \ln(\cosh(Y_N ^m(\theta) - g(X_N ^m)))\Bigg)
\end{multline}

\newpage
\subsection{Convergence Test}
\noindent
In this paper, we run experiments for a large number of epochs to ensure guaranteed convergence for both, DNN and TNN. For defining convergence for loss, we use the criteria in the pseudocode \ref{alg:two} based on smoothed time series of mean loss \cite{Dataaug}.  

\begin{algorithm}[hbt!]
\caption{Convergence Test}\label{alg:two}
\begin{algorithmic}
\State \textbf{Input}: Observed Time Series of Mean Loss for 100 runs over 3000 epochs L=\{$l_1, l_2, \dots, l_{3000}$\}, window size($w$), batch size($b$), threshold($h$), tolerance($t$)\newline\;
\State \textbf{Smoothing}: Initialize $\alpha$ and obtain Smoothed Time Series ($\hat{L}$) with same starting point $l_1$ and hence $\hat{L} = \{\hat{l}_1, \hat{l}_2, \dots, \hat{l}_{3000}\}$ where $\hat{l}_1$ = $l_1$ and $\hat{l}_{i} = \alpha \hat{l}_{i-1} + (1-\alpha)l_i$ for $i>1$ \newline

\For{i=1 to epochs-$w$} \newline

    \State $Window$ = $\hat{L}[i:i+w]$\;
    \State $Diff$ = $\lvert mean(Window[:b])\rvert$ - $\lvert mean(Window[w-b:])\rvert$ \newline
    \Comment {Difference between the loss trajectory at the start and the end of the window for a batch of entries} \newline
    \If{Entries($Window$) $<$ $h$ and $Diff$ $<$ $t$}
        \State return $i$  \newline
        \Comment{Ensure all the elements in window are less than a threshold and $Diff$ is less than a tolerance} 
    \Else
        \State continue 
      \EndIf
        \State epochs 
\EndFor
\end{algorithmic}
\end{algorithm}


\end{document}